\pdfoutput=1
\documentclass{ieeeaccess}
\usepackage{cite}
\usepackage{amsmath,amssymb,amsfonts}
\usepackage{algorithmic}
\usepackage{graphicx}
\usepackage{textcomp}
\usepackage{bm}
\usepackage{subcaption}
\usepackage{upgreek}
\usepackage{comment}
\usepackage{cuted}

\newcommand\scalemath[2]{\scalebox{#1}{\mbox{\ensuremath{\displaystyle #2}}}}
\def\BibTeX{{\rm B\kern-.05em{\sc i\kern-.025em b}\kern-.08em
    T\kern-.1667em\lower.7ex\hbox{E}\kern-.125emX}}
\begin{document}
\history{Date of publication xxxx 00, 0000, date of current version xxxx 00, 0000.}
\doi{10.1109/ACCESS.2017.DOI}

\title{Design of Dynamics Invariant LSTM for Touch Based Human-UAV Interaction Detection}
\author{\uppercase{Anees Peringal $^*$}\authorrefmark{1, 2} ,
\uppercase{Mohamad Chehadeh $^*$}\authorrefmark{1, 2}, \uppercase{Rana Azzam}\authorrefmark{1,2}, \uppercase{Mahmoud Hamandi}\authorrefmark{1,2},  \uppercase{Igor Boiko}\authorrefmark{3}, and
\uppercase{Yahya Zweiri}\authorrefmark{1,2}}
\address[1]{Department of Aerospace Engineering, Khalifa University, Abu Dhabi, United Arab Emirates. }
\address[2]{Khalifa University Center for Autonomous Robotic Systems (KUCARS), Khalifa University, Abu Dhabi, United Arab Emirates.}
\address[3]{Department of Electrical Engineering and Computer Science, Khalifa University, Abu Dhabi, United Arab Emirates. }

\tfootnote{$^*$ Authors contributed equally. This work was funded by the Khalifa University of Science and Technology under Awards  CIRA-2020-082 and RCI-2018-KUCARS. }
\markboth
{Author \headeretal: Preparation of Papers for IEEE TRANSACTIONS and JOURNALS}
{Author \headeretal: Preparation of Papers for IEEE TRANSACTIONS and JOURNALS}

\corresp{Corresponding author: Mahmoud Hamandi (e-mail: mahmoud.hamandi@ku.ac.ae).}

\begin{abstract}
The field of Unmanned Aerial Vehicles (UAVs) has reached a high level of maturity in the last few years. Hence, bringing such platforms from closed labs, to day-to-day interactions with humans is important  for commercialization of UAVs. One particular human-UAV scenario of interest for this paper is the payload handover scheme, where a UAV hands over a payload to a human upon their request.
%\textcolor{red}{Human-robot interaction} is emerging as a fundamental field of research, primarily after the proliferation of robots in populated environments. Thereupon, detecting human interactions on aerial robotic vehicles will lay the foundation for a vast amount of applications, where UAVs may partake in critical social activities. 
In this scope, this paper presents a novel real-time human-UAV interaction detection approach, where Long short-term memory  (LSTM) based neural network is developed to detect state profiles resulting from human interaction dynamics. A novel data pre-processing technique is presented; this technique leverages estimated process parameters of training and testing UAVs to build dynamics invariant testing data. The proposed detection algorithm is lightweight and thus can be deployed in real-time using off the shelf UAV platforms; in addition, it depends solely on inertial and position measurements present on any classical UAV platform. The proposed approach is demonstrated on a payload handover task between multirotor UAVs and humans. Training and testing data were collected using real-time experiments. The detection approach has achieved an accuracy of 96\%, giving no false positives even in the presence of external wind disturbances, and when deployed and tested on two different UAVs.
\end{abstract}

\begin{keywords}
deep learning, LSTM, physical human-robot interaction, self-calibration
\end{keywords}

\titlepgskip=-15pt

\maketitle

\section{Introduction}
\label{sec:introduction}
\PARstart{O}{ver} the past few years, multirotor unmanned aerial vehicles (UAVs) have been used in abundance in a wide range of civilian and military applications, particularly to perform tasks that are burdensome for humans, such as exploration, aerial photography, and search and rescue \cite{8682048}. Recent research efforts, on the other hand, have been directed towards developing robots that can work closely with humans, in a variety of fields like manufacturing, healthcare, and entertainment \cite{veroustraete2015rise, maghazei2021drones}. To that end, there is an imperative need for reliable human-robot interaction algorithms in pursuit of  the successful deployment of robots in close proximity to humans. Robots need to acquire a set of skills, by means of classical or learning approaches, that facilitate interactions with humans without jeopardizing their safety in their common workplace. For instance, object handover is a very critical skill that robots need to get hold of to effectively carry out collaborative tasks with humans \cite{matheson2019human, mainprice2012sharing,cuniato2021hardware}. Although this skill has been widely researched for robotic arms, major efforts are yet required to further develop and advance handover algorithms for aerial mobile robots, the mobility of which makes them ideal for such applications. Nevertheless, UAVs suffer from limitations in size, weight, and power, which raise the need for lightweight algorithms that require no additional sensors onboard the vehicle. 

For humans to effectively perform collaborative tasks with UAVs, it is crucial to establish a channel of communication between them. From the human's side, observing the behaviour, speed and orientation of the UAV might be sufficient if the UAV is trained to convey its actions adequately to the user \cite{khambhaita:hal-01568841,8956408}. On the other hand, UAVs need to be explicitly trained to detect cues specific to the humans' intention to interact. Upon physically interacting with the UAV, humans exert a certain force, which we conjecture to exhibit unique profiles. Such profiles, if accurately detected and identified, may serve as indicators of how the UAV should move forward with its ongoing task. Therefore, this paper presents a novel learning-based interaction states profile (ISP) detection technique to identify human interactions on UAVs. The approach is demonstrated in an object handover scenario, yet is applicable to any other human-UAV interaction application.

A long-short-term memory (LSTM) network is developed and trained to detect arbitrary dynamics and interaction profiles perceived by the UAV upon physical contact with a human. The proposed approach is designed to be platform-agnostic while relying solely on measurements obtained from the proprioceptive sensors onboard the UAV, i.e. IMU and position sensors. After training and deploying on a physical UAV, the proposed ISP detection algorithm resulted in a successful demonstration of a payload handover to humans through physical interaction.

Experimental results have proven the reliability of the proposed ISP detection approach across a wide range of varying profile complexities, where a success rate of 96\% was achieved. The approach was also shown to be robust against random interactions, external wind, and changes in UAV dynamics. In addition, the proposed approach was demonstrated to successfully work on different platforms. Particularly, the ISP detector was trained on a quadrotor and was tested on a vastly different hexarotor. This is attributed to the fact that input to the proposed algorithm is preprocessed to take into account the UAV dynamics and hence resulting in better generalization across different platforms.

\subsection{Related Work}
Detecting an interaction between humans is trivial because a human can obtain information about an interaction both proactively, through vision and audition, and reactively through tactile interactions. Humans also utilize prior experience through their cognitive capabilities to achieve fluid interactions by adequately anticipating and reacting to events and stimuli \cite{Strabala}. When attempting to involve a robotic agent for physical human-robot interaction, a new set of challenges needs to be addressed, due to the inherent limitations in onboard sensors, and the unreliable knowledge of the interaction model. 
Relevant work in the literature can be broadly categorized based on the platform used in the study. Most work on object handover and physical human interaction is done on a robotic arm. In this paper, however, the focus will be devoted to object handover by a UAV.

\subsubsection{Robotic Arm}
Object handover and interaction detection with robotic arms can be classified into proactive and reactive methods. Proactive interaction detection methods get information about the receiver before the actual interaction occurs. These methods make use of vision sensors or motion capture of the human. For example, in \cite{Medina2016} a glove fitted with tactile sensor patches, motion capture markers, and force/torque sensors is worn by the user during the handover. The glove is used for position and force estimation of the human during the handover. The glove is used for experiments in human-human handover to obtain insights into trajectories and forces humans use to hand objects over. Using these insights, a robot-to-human handover algorithm is proposed aimed at fluid handovers. This algorithm is also able to reduce the internal forces that act on the object to protect it from damage during handover. Since this method requires the human to be wearing a specialized precalibrated sensor, it is difficult to generalize to different platforms.

Reactive interaction detection methods make use of sensor data after the interaction has begun. This may include sensors such as force/torque sensors, tactile sensors, etc. In \cite{GmezEguluz2019,Hendrich2014NaturalRH}, tactile sensors are used to find when the receiver is ready to accept the payload. With the help of the above mentioned sensors, the robot was able to reject external perturbations and release the object only when the human interacts in a particular direction. This approach, however, has the drawback that it requires specialized sensors. In some reactive methods, there is no need for specialized sensors. If good knowledge of the robotic arm is available, then the external force acting on the object can be estimated. Such a  method was adopted in \cite{Roveda2021}, where an extended Kalman filter is used to estimate the external forces acting on the robotic arm and implement a force controller for various tasks. Note that in this method, the robot parameters are considered to be known prior to the experiment. On the other hand, the method in \cite{Hanafusa2019} employs a recurrent neural network (RNN) trained to identify the dynamics of a two-link robotic arm, where the network estimates the torques generated by the arm from the knowledge of the joint's angles and their derivatives. The RNN is validated against the measurements from a torque sensor.

\subsubsection{UAV Platform}

Interaction with UAVs introduces new challenges that need to be accounted for as compared to a robotic arm; (1) a UAV platform is inherently unstable, hence the stability of the UAV during interaction has to be taken into account. (2) the size, weight and power of a UAV are limited as compared to fixed manipulators. 

In \cite{Yksel2019}, an onboard force/torque six-degree sensor is used to measure external forces acting on a UAV. A passivity based controller is then used to guarantee the stability of the platform following an external interaction. This approach was demonstrated on a tethered multirotor UAV which does not require an onboard battery. In \cite{Rajappa2017}, an array of force sensors is used along with a disturbance observer to detect human interaction on a UAV platform. The measured force and direction are then used to design an admittance controller. Additional onboard sensors on a UAV will increase the weight of the UAV and thus decrease its performance. Due to the additional cost and weight added by the above sensors, in this work, we will focus solely on sensors available on off-the-shelf UAVs.

In \cite{McKinnon2016}, a sensorless force estimation is done using an unscented Kalman filter with a quaternion-based controller for a UAV to estimate the force and torque acting on the UAV. The authors show that this method can estimate the force and torque profiles in real-time. However, it cannot differentiate between types of forces acting on the UAV. These force estimation methods also require good prior knowledge of the UAV  being used which hinders the transferability to different platforms. In \cite{Tomic2014}, external wrench applied on a UAV platform is estimated based on the proprioceptive sensors onboard the UAV. The estimated external wrench is used to design an admittance and impedance controller for interactions. 

%%%%
The majority of the surveyed research focuses on robotic arms used for payload handover. Interactions with UAVs mandate considering some aspects that are not critical for robotic arms, such as weight. For example, using additional sensors onboard the UAV is not preferable, given its payload constraints. On the other hand, there exist force estimation methods that do not require additional sensors, but they mostly rely on the knowledge of the UAV model, and hence, can only be used for particular platforms. 

To the best of the authors' knowledge, none of the methods proposed in the literature makes use of the profile of states on a UAV for the purpose of differentiating between interactions, which can then be used to encode the communication between a human and a UAV during a collaborative task.

\subsection{Contribution}

Based on the surveyed literature, we propose an AI-based approach for detecting and identifying the profile of the states during a human interaction on a UAV. The proposed approach has a major advantage over existing methods in that the training data and inference of the LSTM are independent of the platform dynamics. Hence, a single model can be used for ISP detection on multiple UAV platforms, and model training can be done using data collected from any UAV platform. In summary, the contributions of this paper are as follows:
\begin{itemize}
    
    \item A novel LSTM-based interaction states profile (ISP) detection approach is proposed to discern human interactions with UAVs, through proprioceptive measurements like IMU and position sensor measurements.
    
    \item A novel data pre-processing technique is developed to make the proposed ISP detection approach invariant to UAV dynamics and hence achieve generality across various platforms without fine-tuning of the trained model. Specifically, the estimated ISP is transformed to a new domain, which we refer to as the training and inference domain (TID), before being processed by the ISP detection approach. The transformation is based on the UAV dynamic parameters identified using the DNN-MRFT approach proposed in \cite{ayyad2020multirotors}.
    
    \item The validity of the proposed ISP detection approach, and its applicability to different UAVs is demonstrated experimentally for different human-UAV interaction scenarios. The detection accuracy on the experimental test set was more than 96\% with no false positives even in the presence of challenging wind conditions, and unwanted random pushes from nearby users. A video summary of the experimental results can be found at \cite{paper_video}.
\end{itemize}

\subsection{Paper Structure}

This paper is structured as follows. Section \ref{model} summarizes the nonlinear modelling of UAV dynamics, the linear decoupled approximations, and the identification of the UAV model parameters. In Section \ref{Section 3}, the pre-processing of the data by a transformation to the TID based on the identified UAV model parameters is presented. In Section \ref{LSTM}, the design and training of the LSTM-based neural network, which operates in the TID is presented. The proposed ISP detection is experimentally verified and an application of this method as payload handover between a robot and a human is demonstrated in real experiments in Section \ref{experiment}. Finally, Section \ref{conclusion} concludes the paper.
%------------------------------------------------
\begin{comment}
\begin{figure*}
    \centering
    \includegraphics[width = 0.9\textwidth, height = 7cm]{images/figure1.png}
    \caption{
    The system is in two different parts, First stage is the system identification stage where the system parameters can be used for designing an optimal controller and to process the data and make the neural network features. The second stage is real-time interaction detection.}
    \label{fig:fig1a}
\end{figure*}
\end{comment}
\begin{figure*}
    \centering
    \includegraphics[width =\textwidth]{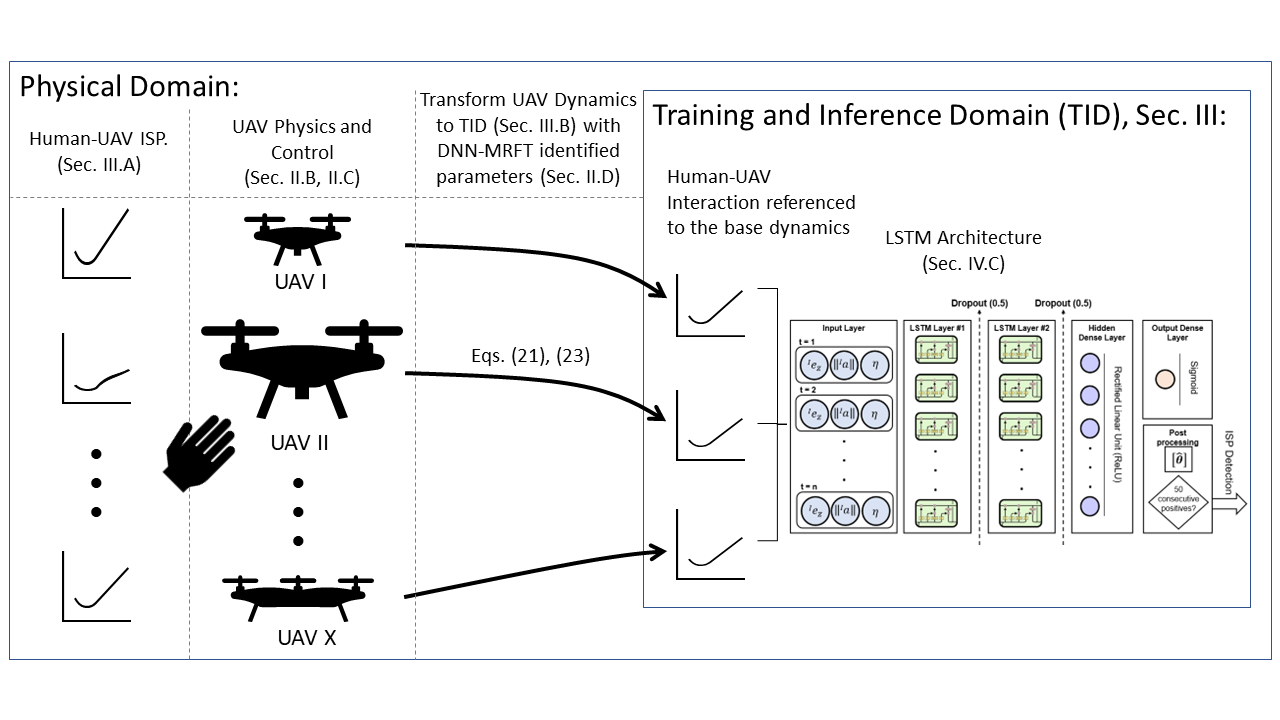}
    \caption{Schematic showing the different constituents of the proposed detection approach.}
    \label{fig:fig1}
\end{figure*}

\section{Modelling, Control and Identification}\label{model}

In this section, the nonlinear model of the quadrotor will be presented, considering the propulsion dynamics and time delays. Then, a linear model that can be used for system identification will be described. This modelling method can be easily extended to any multirotor UAV~\cite{hamandi2021design}, however, in what follows it is derived for a coplanar/collinear platform since a classical quadrotor and a classical hexarotor will be used to demonstrate and verify the proposed approach.

\subsection{Reference Frames and Coordinate System}
\begin{figure}[h]
    \centering
    \includegraphics[scale = 0.5]{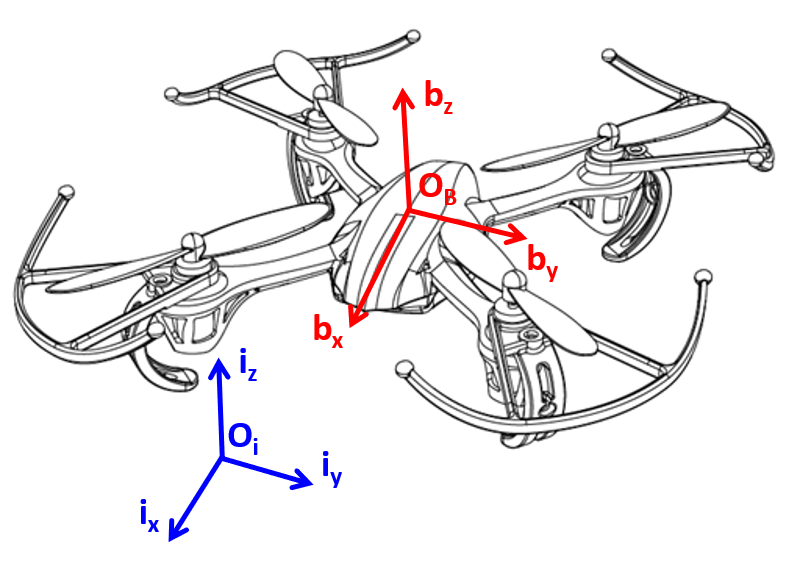}
    \caption{Reference Frames and Coordinate system}
    \label{fig:Coordinate system}
\end{figure}

First, two right-handed reference frames are defined; (1) an Inertial Earth-fixed frame $\mathcal{F}_I$ with axes $\{\bm{i_x}, \bm{i_y}, \bm{i_z}\}$, and (2) a body-fixed frame $\mathcal{F}_B$ with axes $\{\bm{b_x}, \bm{b_y}, \bm{b_z}\}$. The axis $\bm{i_z}$ is chosen to point upwards opposite to gravity, and $\bm{b_z}$ parallel to the direction in which actuators exert thrust. The center of the body-fixed frame is at the center of gravity of the UAV platform and the rotations around the inertial axes are given by Euler angles $\bm{\eta} = [\phi \; \theta \; \psi]$ which represent roll, pitch and yaw. %, and rotations around its $x$, $y$, and $z$ axes are referred to as \(\mathbf{\eta} = [ \phi \; \theta \; \psi]^T\).
For convenience, in what follows a vector may be expressed in a particular reference frame by indexing its components using the frame's symbol. For example, the position vector \({}^I\bm{p}\) is defined with reference to the inertial frame and has the components $[{}^Ip_x \; {}^Ip_y \; {}^Ip_z]^T$. In addition, we use the notion of \(\bar{v}\) to represent a 2D vector obtained from the projection of a 3D vector \(v\) onto the \(x\)-\(y\) defined plane.
The reference frames used in this paper are shown in Fig. \ref{fig:Coordinate system}. 

%The rotation matrix describing the orientation of the hexacopter is defined as ${}_B^IR$ in the SO(3) Lie group; the identity of which is defied as ${}_B^IR{}_I^BR=\mathbf{I}$. The rotational velocity vector is defined as \(\bm{\omega}=[p \; q \; r]^T\), where the Lie algebra \(\bm{\mathfrak{so}}\bm{(3)}\) is set to be around the group's identity \cite{hamel2006attitude}.

%A right-handed, Earth fixed inertial coordinate frame $\mathcal{F}_I$ with basis \( \bm{[i_x, i_y, i_z]}\) where the basis $\bm{i_z}$ is pointing upwards, i.e. opposite to gravity. We also define the body-fixed reference frame $\mathcal{F_B}$ with the basis vector $\bm{b_z}$ parallel to the thrust force generated by the propellers.

\subsection{Non-linear Multirotor model}
Based on the frames of reference defined above, the non-linear model of the UAV will be derived based on the Newton-Euler equations.
First let us denote by $n \in \mathbb{N}$ as the total number of propellers, where $n=4$ for a quadrotor, and $n=6$ for a hexarotor.
The total thrust force \(f_T\), pitching and rolling moments $\tau_{{\phi,\theta}}$, and yawing moments $\tau_{\psi}$ generated by the corresponding propellers, and expressed in $\mathcal{F}_B$ are related to the rotation speed $\Omega$ of each propeller and are described by:
\begin{align}\label{Thrust force}
f_T = \sum_{i=1}^{n}f_i = \sum_{i=1}^{n} k_f\Omega_i^2\\
\tau_{{\phi,\theta}} =  \sum_{i=1}^{n} k_f l_{\phi,\theta}\Omega_i^2 \begin{bmatrix}
sin(\gamma_i)\\
cos(\gamma_i)
\end{bmatrix}\\%\label{rolling and pitching torque}
\tau_{\psi} = \sum_{i=1}^{n}(-1)^{i+1}k_\uptau\Omega_i^2 \label{Yawing torque}
\end{align}
%The pitching and rolling torque $\uptau_{{\phi,\theta}}$ offered by the propellers are given by:
%\begin{equation}\label{rolling and pitching torque}
%\uptau_{{\phi,\theta}} = k_fl_{\phi,\theta}\Omega_i^2
%\end{equation}

where $f_i$ is the thrust produced by the corresponding propeller about its $z$-axis, $k_f$ and $k_\uptau$ are the thrust and drag coefficient of each propeller, $l_{\theta,\phi}$ is the moment arm connecting the platform's center of mass to the each propeller's center of mass, and $\gamma_i$ is the angle between $\bm{b_x}$ and the arm connecting the platform to the corresponding propeller.
%The yawing torque $\uptau_{\psi}$ is produced by the reactive force due to the torque created by the propellers and will be of the form:
%\begin{equation}\label{Yawing torque}
%\uptau_{\psi} = (-1)^{i+1}k_\uptau\Omega_i^2
%\end{equation}
%where $k_\uptau$ is the torque coefficient of the propeller.

The UAV is assumed to be symmetric about $\bm{b_x}$ and $\bm{b_y}$, with its center of gravity coinciding with the origin of $\mathcal{F}_B$. With these assumptions, the product of inertia of the UAV is zero, and therefore, the inertia matrix is $\textbf{J} = diag(J_x, J_y, J_z)$. As such, the Newton-Euler equations for the multirotor UAV are given by:
\begin{equation}\label{eqn:NEE}
\begin{bmatrix}
m I_{3\times3} & 0_{3\times3}\\
0_{3\times3} & \bm{J}
\end{bmatrix}
\begin{bmatrix}
\bm{\dot{V}}\\
\bm{\dot{\omega}}
\end{bmatrix}
=\begin{bmatrix}
{}^B\bm{F}\\
{}^B\bm{\tau}
\end{bmatrix}
\end{equation}

\begin{equation}\label{eqn:ForceB}
{}^B\bm{F} = f_T \bm{b_z} - {}_I^BRmg\bm{i_z} - \bm{\alpha}
\end{equation}

\begin{equation}\label{eqn:TorqueB}
    {}^B\bm{\tau} = \begin{bmatrix}
\tau_\phi \\ \tau_\theta \\ \tau_\psi
\end{bmatrix} - \bm{\lambda}
\end{equation}

\noindent where  $m$ is the mass of the UAV, $\bm{\dot{V}}$ is the acceleration of the UAV, and $\bm{\dot{\omega}}$ is the angular acceleration of the UAV. $\alpha$, and $\lambda$ are arbitrary constants that depend on the translational and rotational drag on the UAV, and captures motion inflow and blade flapping drags on the propellers. If the cross-coupling, observed in (\ref{eqn:NEE}), (\ref{eqn:ForceB}) and (\ref{eqn:TorqueB}) is neglected, the model can be simplified as follows.
\begin{equation}\label{eqn:QCmodel}
\begin{split}
    {}^I\ddot{p}_x &= \frac{1}{m}\left( (c_\psi s_\theta c_\phi + s_\psi s_\phi)f_T -\alpha_x({}^I\dot{p}_x,\eta,\Omega)  \right),  \\
    {}^I\ddot{p}_y &= \frac{1}{m} \left( (s_\psi s_\theta c_\phi - c_\psi s_\phi)f_T - \alpha_y({}^I\dot{p}_y,\eta,\Omega) \right)  ,  \\
    {}^I\ddot{p}_z &= \frac{1}{m} \left( c_\phi c_\theta f_T - g - \alpha_z({}^I\dot{p}_z,\eta,\Omega) \right) ,  \\
    \ddot{\phi} &= \frac{1}{J_x} \left( \tau_\phi - \lambda_\phi(\omega,\Omega) \right), \\
    \ddot{\theta} &= \frac{1}{J_y} \left( \tau_\theta - \lambda_\theta(\omega,\Omega) \right), \\
    \ddot{\psi} &= \frac{1}{J_z} \left( \tau_\psi - \lambda_\psi(\omega,\Omega) \right). \\
\end{split}
\end{equation}

%Propulsion dynamics are considered in the model where
A brushless DC motor (BLDC) is used for propulsion, controlled with an electronic speed controller (ESC). Each ESC receives the desired propeller rotational speed $\Omega_i$, and controls the BLDC to achieve the desired command $u_i$, assumed to be proportional to $\Omega_i^2$.
%While the 
%receives a command $u_i$ and produces a propeller squared rotation of $\Omega_i^2$. 
The relationship between the desired command applied by the ESC and the produced thrust or torque by the corresponding BLDC is approximated by a first-order plus time delay model \cite{Cheron2010}, and is given by:
\begin{equation}\label{actuator_dynamics}
\dot{f_i}(t)T_p + f_i(t) = K_p u_i(t-\tau_p)
\end{equation}
where \(T_p\) is the propulsion time constant, \(K_p\) is the propulsion dynamics gain, \(\tau_p\) is the propulsion dynamics delay and $t$ is the time variable.
\subsection{Linearized Inner Loop Dynamics}
The inner loop dynamics of a multirotor UAV consists of the altitude dynamics, the attitude (i.e. roll and pitch) dynamics, and the yaw dynamics. Both altitude and attitude dynamics of a multirotor UAV have the same model structure, yet the model parameters differ for each. For example, the model for the pitch loop is given by:
\begin{equation}\label{attitude_transfer_function}
%G_{att,alt} = \frac{K_{eq}}{s(T_1s+1)}
T_1\ddot{\theta}(t) +\dot{\theta}(t) = K\tau_\theta  
\end{equation}
where \(T_1\) represents drag time constant, and \(K\) represents the loop gain. The pitch state variable \(\theta\) might be simply substituted by the state variables \(\phi\), and \({}^Ip_z\) to obtain the other models. Yet, it is required to substitute \(\tau_\theta\) in \eqref{attitude_transfer_function} with the propulsion dynamics presented in \eqref{actuator_dynamics}. The full pitch dynamics are given by
\begin{equation}\label{inner_loop_tf}
%G_{inner} = \frac{k_{eq}e^{-\tau_{in} s}}{s(T_1s+1)(T_{prop}s+1) }
T_1T_p\dddot{\theta}(t) +(T_1+T_p)\ddot{\theta}(t) +\dot{\theta}(t) = K_{eq}u_\theta(t-\tau_\theta)
\end{equation}
where \(\tau_\theta\) includes the total loop delay, i.e. due to propulsion, sensors, and digital circuits. 

The yaw dynamics has a different model structure compared to the other inner loop dynamics of a multirotor UAV and can be modelled as a second-order system \cite{ayyad2020multirotors}:
\begin{equation}\label{yaw dynamics}
%G_{yaw} = \frac{K_{eq}e^{-\tau s}}{s(T_{prop}s +1)}
T_p\ddot{\psi}(t) +\dot{\psi}(t) = K_{eq}u_\psi(t-\tau_\psi)
\end{equation}

Note that the propulsion dynamics are shared among all inner loops. Also, the drag time constant \(T_1\) and the equivalent gain \(K_{eq}\) would be different for every inner loop.

\subsection{Model Parameters Identification}\label{system-identification}
The unknown altitude and attitude models parameters can be identified in real-time using the recently developed approach of DNN-MRFT \cite{Ayyad2020}.  This method required decoupling of the UAV dynamics into SISO systems, as done in the previous section. Let the vector $\bm{d} = [K_{eq} \, T_1 \, T_p \, \uptau]\in D$ represent the model parameters of a given altitude or attitude loop. The model parameters' bounds provided in \(D\) are chosen such that all UAV designs of interest fall within. The DNN provides a map between a test signal produced experimentally, and the unknown parameters vector \(\bm{d}\). The modified relay feedback test (MRFT) \cite{Boiko2012} is used to produce the system response, which is given by:
\begin{equation}\label{eq_mrft_algorithm}
u_M(t)=\scalemath{0.8}{
\left\{
\begin{array}[r]{l l}
h\; &:\; e(t) \geq b_1\; \lor\; (e(t) > -b_2 \;\land\; u_M(t-) = \;\;\, h)\\
-h\; &:\; e(t) \leq -b_2 \;\lor\; (e(t) < b_1 \;\land\; u_M(t-) = -h)
\end{array}
\right.}
\end{equation}
\noindent where $b_1 = -\beta e_{min}$ and $b_2 = \beta e_{max}$. $e_{max}$ and $e_{min}$ are the maximum and minimum error signal values when the system undergoes stable oscillations. $\beta$ is a tunable parameter which determines the phase of the excited oscillations as, $\varphi = \arcsin{\beta}$. In this paper, $\beta$ is chosen as the global optimal $\beta$ that minimizes ISE error for a step test as described in \cite{Chehadeh2019, Boiko2014}. The MRFT always produces stable oscillations for the case of inner loops of multirotor UAVs as was shown in \cite{ayyad2020multirotors}. 

A classification DNN would provide more advantages compared to a regression DNN. First, the discretization of DNN output increases the efficiency of training of the DNN with a custom loss function that depends on optimal control parameters \cite{Ayyad2020}. Second, a classification DNN would allow us to obtain optimal controllers of the UAV in real-time based on $\bm{d}\in \bar{D}$, where \(\bar{D}\) represents a discretization of \(D\) and \(|\bar{D}|\) is the number of the output classes of the DNN. Based on the simulated MRFT responses of all systems in \(\bar{D}\), the DNN would select an element from $\bar{D}$ that best describes the system under test.

For a process $G_i(s)$ with model parameters $\bm{d}_i$, there exists an optimal controller $C_i(s)$ which minimizes a particular cost functional. $\bar{D}$ is discretized in such a way that the percentage change in the integrated square error (ISE) when the controller $C_i(s)$ is used with an adjacent process \(G_j\) is less than some predefined value $J^*$ (in this paper we chose $J^* < 10\%$ as in \cite{Ayyad2020}). This can be formulated using the relative sensitivity function, which indicates the robustness of the system to the changes in process parameters and is governed by the following equation \cite{rohrer1965sensitivity}:
\begin{equation}\label{eq_performance_deterioration}
J_{ij} = \frac{Q(C_i, G_j) - Q(C_j, G_j)}{Q(C_j, G_j)} \times 100 \%
\end{equation}
where \(J_{ij}\) represents the degradation in performance due to applying controller \(C_i\), which is the optimal controller for the process \(G_i\) and a sub-optimal controller for the process \(G_j\). \(Q\) denotes the integral square error (ISE) of the step response of the closed loop system:
\begin{equation}
    Q(G(s),C(s)) = \frac{1}{T_s}\int_0^{T_s} e(G,C)^2dt 
\end{equation}

\section{Design of Training and Inference Domain}\label{Section 3}
To address the shortcoming of platform specific AI, we introduce the concept of a physical domain at which the UAV dynamics are characterized, and a TID at which the AI model is trained and the real-time inference occurs. For this to be achievable, the AI in TID has to be invariant to the specific UAV dynamics. Fig. \ref{fig:fig1} shows how a human-UAV ISP can be transformed from the physical domain to the TID using the DNN-MRFT identified model parameters. In this work, we have designed three different types of human-UAV interactions to be detected by three different LSTM models. These interactions are:
\begin{enumerate}
    \item Single downward pull (SDP) of the UAV.
    \item Consecutive double downward pulls (CDDP) of the UAV. 
    \item Single yawing twist (SYT) of the UAV.
\end{enumerate}
\begin{figure}
    \centering
    \includegraphics[scale = 0.6]{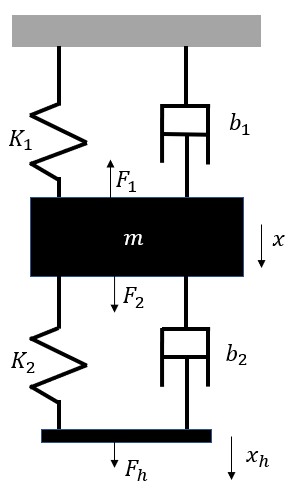}
    \caption{Spring mass damper analogy of the interaction}
    \label{fig:smd}
\end{figure}
A spring-mass damper system, depicted in Fig.\ref{fig:smd}, can be used as an analogy for the analysis of the human-UAV interaction. In this analogy, the mass corresponds to the UAV, the first set of spring-damper corresponds to the closed loop UAV system, and the second set of spring-damper corresponds to the dynamics of the human arm. Human dynamics have been modelled as a spring-mass damper system in the literature \cite{human_state_aware}. The force exerted by the human is $F_h$.
The spring mass damper system is described by:
\begin{align}
    m\ddot{x} &= F_1 -F_2\\
    F_1 &= -K_1(x) -b_1\dot{x} \\
    F_2 &= K_2(x-x_h) +b_2(\dot{x} - \dot{x_h})
\end{align}
Comparing this system to the real human-UAV interaction system, the stiffness $K_2$ and damping $b_2$ terms of the human arm cannot be identified. Similarly, while the position $x$ and acceleration $\ddot{x}$  of the UAV are measured, while the position of the human arm $x_h$ is not. The proposed learning-based method should implicitly estimate the force of the human hand $F_h$ given the measurable parameters. As the human dynamics are unobservable, the LSTM should be robust enough that differences in $K_2$ and $b_2$ between different humans do not affect its output. This is done by including training data obtained from tests with different humans. This makes human dynamics part of the TID and hence, the LSTM effort is devoted to learning the human-invariant interaction behavior.

\begin{figure}
    \centering
    \includegraphics[width = 0.45\textwidth]{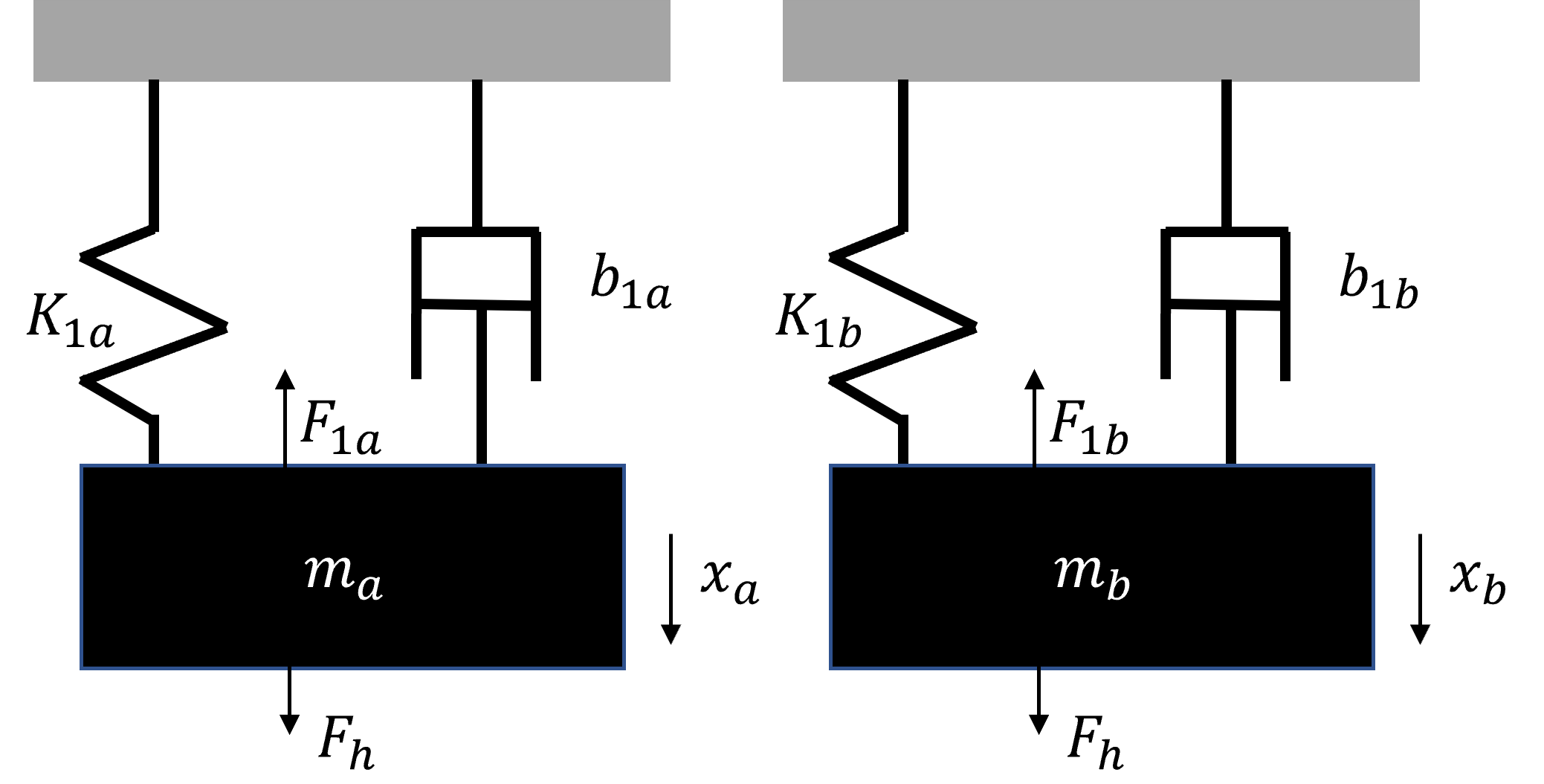}
    \caption{Two spring mass damper systems to compare the response to the same input force}
    \label{fig:two_rigid}
\end{figure}

It should be noted that using the spring-damper analogy, the LSTM would overfit the dynamics characterized by \(K_1\), and \(b_1\). To overcome this limitation, we assume that the human exerted force profile is the same regardless of the UAV in use. To verify this assumption, we have conducted an experimental test where we have asked seven human subjects to detach a payload carried by a UAV. To make the handover maneuver as natural as possible, we used a cup of water as a payload and the subjects were not provided with any instructions apart from asking them to receive the payload by pulling it downwards. We recorded the estimated $F_h$ of the first downward pull of the seven human subjects, and it was seen that the peak force differs only by 10\% between all the subjects. As such, if $F_h$ is assumed to be similar in all platforms, a relation between the responses of different spring mass damper UAV systems can be obtained. Under this assumption, consider two new spring-mass damper systems representing UAV (a) and UAV (b), with the same applied \(F_h\), as in Fig.\ref{fig:two_rigid}. The response to an external force $F_h$ is modelled by:
\begin{equation}\label{eq_newton_two_mass}
    \begin{aligned}
    m_a\ddot{x_a} &= -K_{1a}x_a - b_{1a}\dot{x_a} +F_h \\
    m_b\ddot{x_b} &= -K_{1b}x_a - b_{1b}\dot{x_b} +F_h
\end{aligned}
\end{equation}

equating the two forces gives the relation (written in the Laplace domain for convenience):
\begin{equation}\label{eq_uav_dynamics_transformation}
       \frac{X_a(s)}{X_b(s)} = \frac{m_bs^2 +b_{1b}s +k_{1b}}{m_as^2 +b_{1a}s +K_{1a}} 
\end{equation}
so if \(x_a\) is selected as a feature for LSTM training and all training happens based on the UAV platform (a), \eqref{eq_uav_dynamics_transformation} can be used to transform data collected from UAV platform (b). In this case, the UAV platform (a) resides in the TID, and its dynamics are referred to as the \textit{base} dynamics. UAV platform (b) resides in the physical domain, which might contain an arbitrary number of different UAV dynamics. In the rest of this section, we discuss the aspects of the selection of features for the considered interactions, i.e. SDP, CDDP, and SYT, and we derive transfer functions for the transformation of the proposed features from the physical domain to the TID.

\subsection{Feature Selection}\label{formulation}

\begin{figure}
    \centering
    \includegraphics[width = 0.45\textwidth]{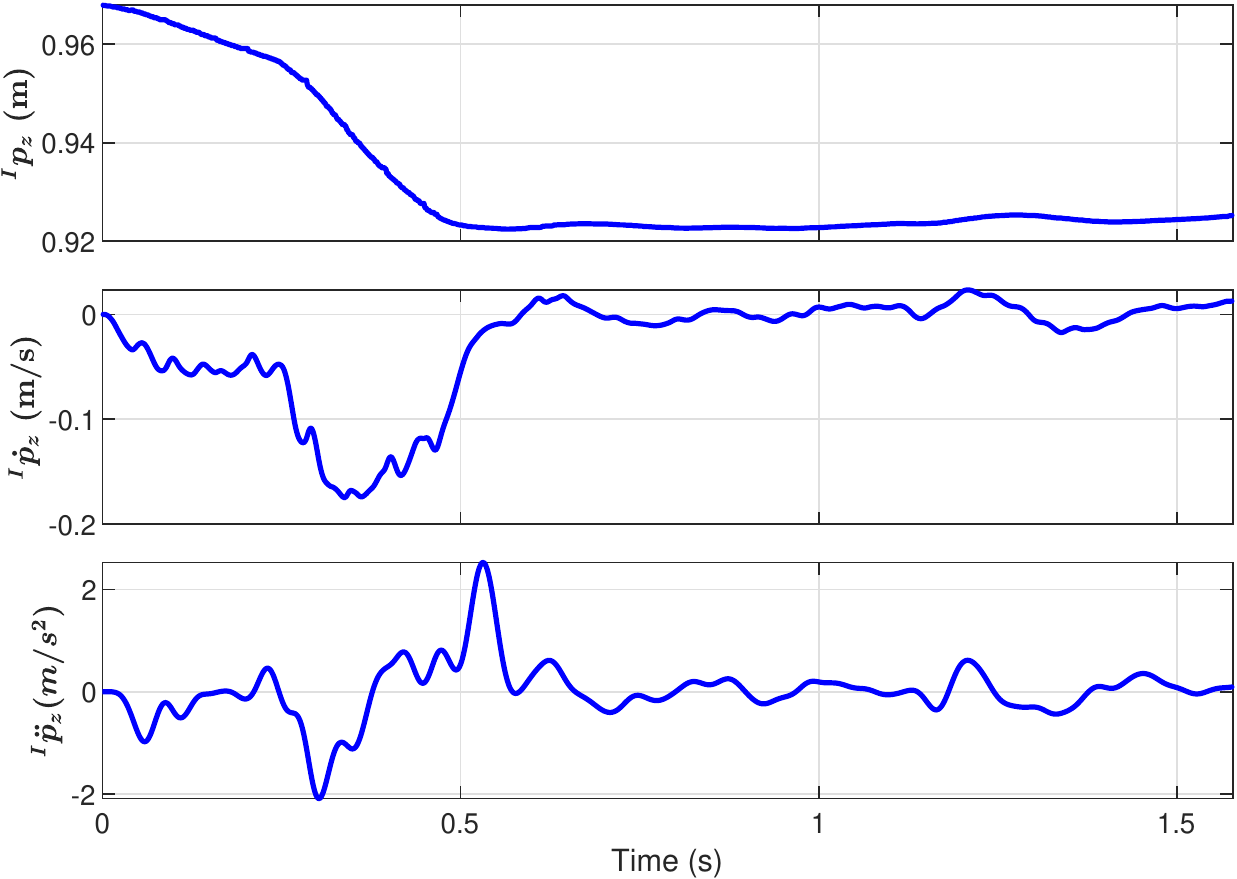}
    \caption{Experimentally obtained position, velocity and acceleration during a single downward pull interaction.}
    \label{fig:my_label}
\end{figure}

To avoid the installation of additional sensors onboard a multirotor UAV, the interaction detection LSTM must depend on observable states measured by the UAV avionics. These selected states are called features of the ISP, from which the human-UAV interaction is detectable by the LSTM. Obviously, the features that are selected will depend on the type of interaction that needs to be detected. The states that can be estimated from a UAV platform fitted with IMU and a position and heading sensors is tabulated in Table \ref{tab:Propioceptive}. Note that Table \ref{tab:Propioceptive} assumes the use of a motion capture (Mocap) system, which can be replaced by other positioning and heading measurement sources. 

We have performed initial testing for the SDP, CDDP, and SYT interactions to select suitable states that would define the ISP for each. The features selected for the SDP, and CDDP interactions are shown in Table \ref{tab:features}. The selected features for the SYT interaction detection LSTM are the following states: pitch \(\theta\), roll \(\phi\), yaw \(\psi\), body accelerations \([{}^Ba_x\;{}^Ba_y\;{}^Ba_z]^T\), and controller output of the yaw loop.

\begin{table}[]
    \centering
    \def\arraystretch{1.4}
    \begin{tabular}{|c|c|c|}
        \hline
        Estimated State & Sensor & Notes \\
        \hline
        ${}^I_BR$, $\eta$ & IMU and Mocap & Mocap for yaw  \\
        \hline
        ${}^B\bm{\omega}$ & Gyroscope & \\
        \hline
        ${}^I\bm{p}$ & Mocap & \\
        \hline
        ${}^I\bm{\dot{p}}$ & Mocap, and accelerometer & Sensor fusion \\
        \hline
        ${}^I\bm{\ddot{p}}$ & Accelerometer & Uses ${}^I_BR$ to remove \({}^I\bm{g}\)\\
        \hline
    \end{tabular}
    \caption{The state measurements that can be obtained for low-cost UAVs. In outdoor settings, the Mocap can be replaced by a combination of GPS, barometer, and magnetometer.}
    \label{tab:Propioceptive}
\end{table}

{\renewcommand{\arraystretch}{2}
\begin{table}[]
    \centering
    \bgroup
    \def\arraystretch{2.5}
    \begin{tabular}{|c|c|}
    \hline
        Feature & Equation \\\hline
        ${}^Ie_z$ & ${}^Ip_{z_{ref}}-{}^Ip_z$ \\ \hline
        $\|{}^I\bm{a}\|$ & $ \sqrt{{}^Ia_x^2 +{}^Ia_y^2+{}^Ia_z^2}$\\ \hline
        $\eta$ & \(\arctan{(\frac{\sqrt{({}^B_IR\bm{i}_z\cdot\bm{i}_x)^2+({}^B_IR\bm{i}_z\cdot\bm{i}_y)^2}}{{}^B_IR\bm{i}_z\cdot\bm{i}_z})}\) \\ \hline%$\sqrt{{}^\mathcal{B}_\mathcal{I}R{\bm{i_{z_x}}^2} + {}^\mathcal{B}_\mathcal{I}R{\bm{i_{z_y}}^2}}$\\ \hline
        %$\Delta \Psi$ & $\tan^{-1}\left(\frac{{}^I\mathbf{a}_y}{{}^I\mathbf{a}_x}\right) - \tan^{-1}\left(\frac{{}^B\mathbf{i_z}_y}{{}^B\mathbf{i_z}_x}\right)$ \\ \hline
    \end{tabular}
    \egroup
    \caption{LSTM features selected to represent the ISP for SDP and CDDP interactions. The state \(\eta\) represents the tilt angle of \(\bm{b}_z\) with respect to \(\bm{i}_z\).}
    \label{tab:features}
\end{table}
}

\subsection{Dynamics Transformation to the Training and Inference Domain}\label{dynamics_invariance}
The UAV specific dynamics are required to define the transformation to the TID. The UAV inner loop dynamic parameters are assumed to be available through DNN-MRFT as described in Section \ref{system-identification}.

For the \({}^Ie_z\) feature, the altitude loop dynamics have to be considered, with the assumption that \(\bm{b_z}\) and \(\bm{i_z}\) are almost aligned (note that it is unsafe to interact with the drone otherwise). By proceeding with the assumption that $F_h$ acting on two different UAVs are the same as stated in \eqref{eq_newton_two_mass}, we can write the following relation between the errors on two different UAVs differentiated by the indices 1 and 2:
\begin{align*}
    {}^Ie_{z1}\frac{\left(K_{c1}+sK_{d1}\right)K_{p1} +m_1s^2(T_{p1}s+1)}{T_{p1}s+1} =  \\
    {}^Ie_{z2}\frac{\left(K_{c2}+sK_{d2}\right)K_{p2} +m_2s^2(T_{p2}s+1)}{T_{p2}s+1} 
\end{align*}
where \(K_c\) represents the proportional controller gain, \(K_d\) represents the derivative controller gain, \(K_p\) and \(T_p\) are the propulsion system parameters from \eqref{actuator_dynamics}. Note that we assumed that the drag force and the delay in propulsion to be negligible during the interaction (i.e. \(T_1=0\)). The above relation can be rearranged in a transfer function form to give:
\begin{strip}
\begin{equation}\label{eq_tf_four_ordr}
    \frac{{}^Ie_{z_1}(s)}{{}^Ie_{z_2}(s)} = \scalemath{0.8}{\frac{K_{c_2}K_{p_2}+K_{p_2}(T_{p_1}K_{c_2}+K_{d_2})s+(T_{p_1}K_{d_2}K_{p_2}+m_2)s^2+m_2(T_{p_1}+T_{p_2})s^3+m_2T_{p_1}T_{p_2}s^4}{K_{c_1}K_{p_1}+K_{p_1}(T_{p_2}K_{c_1}+K_{d_1})s+(T_{p_2}K_{d_1}K_{p_1}+m_1)s^2+m_1(T_{p_2}+T_{p_1})s^3+m_1T_{p_2}T_{p_1}s^4}}
\end{equation}
\end{strip}

When fully defined, the transfer function in \eqref{eq_tf_four_ordr} provides a transformation of the \({}^Ie_{z_1}\) of UAV 2 in the physical domain, to the base dynamics in the TID. But realizing \eqref{eq_tf_four_ordr} requires two considerations. First, the states usually available on commercial multirotor UAVs do not include estimates for \({}^Ip_z^{(3)}\) and \({}^Ip_z^{(4)}\). We assume the ratio of these higher order terms is close to unity, and hence we rewrite \eqref{eq_tf_four_ordr} to obtain the simplified transformation:
\begin{equation}\label{eq_tf_scnd_ordr}
    \frac{{}^Ie_{z_1}(s)}{{}^Ie_{z_2}(s)} = \scalemath{0.8}{\frac{K_{c_2}K_{p_2}+K_{p_2}(T_{p_1}K_{c_2}+K_{d_2})s+(T_{p_1}K_{d_2}K_{p_2}+m_2)s^2}{K_{c_1}K_{p_1}+K_{p_1}(T_{p_2}K_{c_1}+K_{d_1})s+(T_{p_2}K_{d_1}K_{p_1}+m_1)s^2}}
\end{equation}
which is easy to realize due to the presence of \({}^I\dot{p}_z\) and \({}^I\ddot{p}_z\) estimates (refer to Table \ref{tab:Propioceptive}). The denominator in \eqref{eq_tf_scnd_ordr} acts as a filter for the noisy \({}^I\ddot{p}_z\) measurements, which provides advantages for practical real-time realizations. The transformation of the feature \({}^I\dot{p}_z\) can be obtained by neglecting the polynomial terms corresponding to position form \eqref{eq_tf_scnd_ordr}, and similarly, a transformation for the acceleration feature \({}^Ia_z\equiv{}^I\ddot{p}_z\) can be obtained. Note that for the feature \(\|{}^I\bm{a}\|\) in Table \ref{tab:features} the same transformation as of \({}^Ia_z\) is assumed, which is a valid choice due to the UAV underactuated nature.

The second consideration for the practical realization of \eqref{eq_tf_scnd_ordr} is the availability of all the transformation parameters. The DNN-MRFT identifies the time parameter \(T_p\) and the equivalent loop gain \(K_{eq}\) as defined in \eqref{inner_loop_tf}, where \(K_{eq}=K_p/m\), and also provides controller parameters \(K_c\) and \(K_d\). But \eqref{eq_tf_scnd_ordr} requires the knowledge of both parameters \(K_p\) and \(m\), so that one of these must be found prior to the interaction through lab tests. Obtaining the static gain parameters \(K_p\) and \(m\) using lab experiments is straightforward and is much easier compared to the other dynamic parameters present in \eqref{eq_tf_scnd_ordr} which are obtained through the DNN-MRFT. Thus, the DNN-MRFT provides an important automated step in the design of these dynamic transformations.

A similar transformation applied to altitude can be obtained for the angular dynamics. From Table \ref{tab:features}, we have selected the tilt angle \(\eta\) as a feature which we assume to have the same dynamics as the pitch angle presented in \eqref{inner_loop_tf}. This assumption is valid for symmetric multirotor UAVs. Thus, in analogy to the altitude dynamics, the transformation for \(\eta\) dynamics is given by:
\begin{equation}\label{eq_tf_scnd_ordr_ptich}
    \scalemath{0.8}{\frac{{}^Ie_{\eta_1}(s)}{{}^Ie_{\eta_2}(s)} = \frac{K_{c_2}K_{p_2}+K_{p_2}(T_{p_1}K_{c_2}+K_{d_2})s+(T_{p_1}K_{d_2}K_{p_2}+I_{y_2}/l_{y_2})s^2}{K_{c_1}K_{p_1}+K_{p_1}(T_{p_2}K_{c_1}+K_{d_1})s+(T_{p_2}K_{d_1}K_{p_1}+I_{y_1}/l_{y_1})s^2}}
\end{equation}
where \(I_{xy}\) corresponds to the rotational inertia of the tilt dynamics, and \(l_{xy}\) is the equivalent motor to center distance projected on \(\bm{b_x}\times\bm{b_y}\) which is assumed to be constant due to multirotor UAV symmetric design. Unlike the altitude case, a measurement for angular acceleration is not available. Thus, we truncate the transformation in \eqref{eq_tf_scnd_ordr_ptich} to be a first order transformation:
\begin{equation}\label{eq_tf_frst_ordr_ptich}
    \frac{{}^Ie_{\eta_1}(s)}{{}^Ie_{\eta_2}(s)} = \scalemath{1}{\frac{K_{c_2}K_{p_2}+K_{p_2}(T_{p_1}K_{c_2}+K_{d_2})s}{K_{c_1}K_{p_1}+K_{p_1}(T_{p_2}K_{c_1}+K_{d_1})s}}
\end{equation}
which is independent of \(I_{xy}\) and \(l_{xy}\).

Another point to consider when transferring ISPs from the physical domain to the TID is the difference in the sampling rate used in both domains. The sampling rate is adjusted appropriately, either through downsampling, or by upsampling with linear interpolation. The LSTM is trained based on the sampling rate of the TID, which was chosen to be 1 kHz.

\section{LSTM Design}\label{LSTM}
%(Here I will talk about what was the parameter search that I had done on the system and what is the final neural network that I decided to use. When mentioning the parameter search, I should also include a table that summarizes the parameter space. 
%I should also mention the normalizing that happens to the data before it is sent to the LSTM)

\subsection{Architecture} % should this be \paragraph instead of \suvsection?
Long short-term memory (LSTM) neural network is a special type of recurrent neural network (RNN) that is used in various applications with time-series data such as anomaly detection \cite{timeseries} and trajectory prediction \cite{LSTM_traj}. They have the ability to store the previous data input in an internal state of the memory unit which mitigates the vanishing gradient problem faced by RNNs when training on long sequences of data. The memory unit in the network keeps an internal state using the gating mechanism \cite{LSTM_paper}.

%We use an LSTM based machine learning model to detect the ISP acting on the UAV. 

\begin{figure}[h]
    \centering
    \includegraphics[width = 0.5\textwidth]{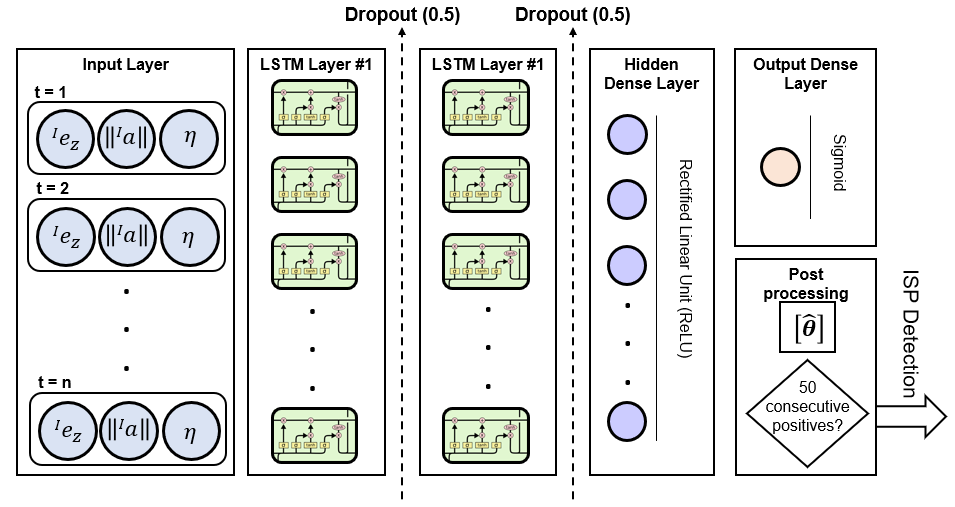}
    \caption{Proposed neural network architecture}
    \label{fig:LSTM Model}
\end{figure}

To determine the optimal neural network architecture for the problem at hand, an automated hyperparameter space searching technique was used. The search space included networks of varying depth, width, activation functions, and combinations of the features in Table \ref{tab:features}, and was trained using two different optimizers, as listed in Table \ref{table:search space}. A total of 55 different neural network structures were evaluated and the best-performing structure, depicted in Fig.\ref{fig:LSTM Model}, was selected based on the prediction accuracy, where it achieved 98.3\% accuracy on the testing dataset.

The adopted neural network structure consists of two LSTM layers with 200 and 100 units respectively, followed by a dense layer whose neurons are activated using the rectified linear unit (ReLU). The output layer consists of a single neuron activated using sigmoid. A classification threshold of 0.5 is used on the output neuron which is represented by a round-off function in Fig. \ref{fig:LSTM Model}. Dropout is used between the hidden layers to regularize the neural network and to avoid overfitting \cite{dropout}. 
%We use the root mean square propagation algorithm for training the neural network. We choose the best performing neural network out of 55 different networks. The search space is shown in Table .
\begin{table}[!h]
\centering
\begin{tabular}{|c|c|}
\hline
\textbf{Parameter} & \textbf{Search Space}  \\ \hline
                number of layers & {1,2,3} \\ \hline
                neurons per layer & {100,200,300} \\ \hline
                activation function & ReLU \\ \hline
                Optimizer & {rmsprop, ADAM} \\ \hline
                features & {2,3,4,6} \\ \hline
\end{tabular}
\caption{Search space for the trained LSTM model}
\label{table:search space}
\end{table}

%Based on the accuracy of the models trained from the automated parameter search, the architecture of the best model is used. This model has an accuracy of the 98.3\%. 
%After the hyperparameter search, we can choose the best neural network based on the prediction accuracy on the testing dataset. The architecture of the neural network that was chosen is shown in (figure \ref{fig:LSTM Model}).

\subsection{Data Generation and Training}
%% new after the comments 
The training data for the neural network is generated from real experiments that are conducted on a UAV. This approach exposes the neural network to training data that closely resemble real-life scenarios. For further generalization, the training data is augmented by introducing noise. A bias of 0.05 and Gaussian noise with zero mean and a standard deviation of 0.0025 is introduced and appended to the original data. This provides a larger data set for training.

The neural network is required to identify human interaction and reject other disturbances that it might encounter during regular operation, such as wind or collisions. Hence, the training dataset included samples of Human-UAV interactions under wind disturbances and samples of random collisions with the UAV.

Collected training data is divided into sequences, the length of which is selected based on the ISP of interest. For instance, the profile resulting from a single pull, and hence the sequence length, is shorter than the profile resulting from a double pull. This will ensure that the desired profile lies within the input sequence to the neural network which will guarantee correct prediction. 
The order of measurements in the sequence will remain unchanged to maintain the characteristic of the profile over time. A total of 14652 sequences were collected, 50\% of which were used for training, 25\% for validation, and 25\% for testing. Prediction accuracy on the testing set will be used for model evaluation.

\subsection{Real-time Inference}

During Real-time inference stage, a stateful LSTM, which gives us the ability to reset the internal states according to our requirement, is used. The internal states of the LSTM are reset when the UAV has taken off and is hovering, awaiting human interaction.

We employ a two-stage discriminator for the neural network. Firstly, the output of the LSTM is a probability distribution \(\mathcal{P}\in[0,1]\). We chose an output  \(\mathcal{P}>0.5\) to represent a positive detection. Secondly, the LSTM model provides a detection result for every time step, however, outlier positive detections are rejected by accepting positive signals that are sustained for a predefined number of time steps, chosen to be 50 ms. This conservative approach is preferred due to 
the nature of the application at hand, where it is preferred not to release the payload unless a receptor human is ready to receive it. To measure the accuracy of the system, a true positive is defined when at least one positive output is present during the interaction, and a false positive when there is a positive output without an interaction.

\section{Experimental Results }\label{experiment}

The proposed LSTM-based ISP detection approach was extensively tested in experimentation using different UAV platforms across multiple scenarios. More specifically, the trained ISP detection model was tested on a quadcopter and a hexacopter while hovering, while moving, under wind disturbance, and using different human force signatures. This section provides, in detail, a description of the experimental setup and an analysis of the obtained results.

\subsection {Experimental Setup}
\begin{figure}
    \centering
    \includegraphics[width =0.38\textwidth]{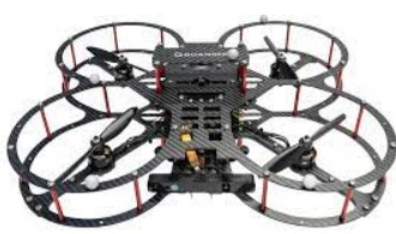}
        \caption{Qdrone}
        \label{fig:Qdrone}
\end{figure}

Two different UAVs are used to verify the applicability of the proposed approach and to demonstrate its generality across various platforms. A quadcopter, which we refer to as the base UAV, was used for collecting training data and for initial verification, and a hexacopter, which we refer to as the testing UAV, was used for verifying the generality of the proposed approach. The specifications of the base and testing UAVs are provided below.
\begin{itemize}

\item{\textbf{Base UAV: }}
The base UAV is a Quanser Qdrone quadcopter, that comes with ducted propellers (Fig. \ref{fig:Qdrone}) which makes it suitable for physical Human-UAV interaction. The Qdrone is a small, 1000g, quadrotor, equipped with an Intel Aero board and a BMI160 onboard IMU. A proprietary \textit{Matlab/Simulink} interface is used to communicate with the drone. A motion capture system is used to measure the position of the UAV at 120Hz.

\item{\textbf{Testing UAV:}}
The DJI F550 Hexarotor UAV is used to test all the neural networks that were trained on the base UAV.
%If the dynamics of the testing UAV is vastly different from the base UAV, and the detection algorithm trained on the base UAV works reliably on the testing UAV, it can be said that the algorithm is indeed Dynamics invariant.
The testing UAV is a 2260g hexarotor that uses the NAVIO2 flight controller hat with a Raspberry pi3B+, where a custom flight control software is run. Xsens 610 is used as an onboard IMU. The LSTM network runs on an onboard intel NUC, since running it on the same RPi3B+ caused major delays in the system. The robot operating system (ROS) is used for communication among the flight controller, the LSTM network, and the motion capture system.   
\end{itemize}

%We run human interaction experiments on the Base UAV (fig:\ref{fig:Qdrone}) because its ducted propellers are safer and would cause less stress on the users \cite{Abtahi2017}. The Validation UAV is only used to show that the LSTM network is able to reliably detect human interaction on different UAVs. It is trivial to see that other experiments like wind disturbance can be extended to the validation system if the need arises.

The experimentation for this research is set up as follows: firstly, experiments are conducted to show that the LSTM network can reliably detect a human interaction on the base UAV it was trained on. Then the generalization of the trained LSTM to different UAVs is demonstrated with the testing UAV.  Finally, an application for the system in the form of payload delivery is then demonstrated.
%In the experimentation, single downward pulls and consecutive double downward pulls are considered. While most experiments are done considering single downward pull,  consecutive double downward pull is included to test the ability of the proposed approach to detect more complex interactions. 

System identification of the two platforms is performed as mentioned in \ref{system-identification}; the resulting parameters are shown in Table.~\ref{tab:hexa_params}. Knowledge of the model for both platforms is necessary when using the ISP detection on a different platform as indicated in \eqref{eq_tf_four_ordr} and \eqref{eq_tf_scnd_ordr_ptich}. The parameters of the altitude channel is tabulated in Table \ref{tab:hexa_params}. 

\begin{table}[h]
    \centering
    \resizebox{0.5\textwidth}{!}{
    \begin{tabular}{|c|c|c|c|c|c|c|}
        \hline UAV & Kp &Kd &K&$T_p$&$T_1$&$\tau$ \\ \hline
         Base UAV & 75 & 13 & 0.1415 & 0.0224 & 0.2776 & 0.0656 \\ \hline
          Testing UAV& 24.809 & 7.4476 & 0.5090 & 0.3 & 0.2 & 0.0128\\ \hline
    \end{tabular}
    }
    \caption{Identification of the base and test UAVs in the altitude loop}
    \label{tab:hexa_params}
\end{table}

%It must be noted that in the base UAV, $T_1 < T_2$ but in the validation UAV, $T_1>T_2$. This is because, if $T_1$ and $T_2$ were interchanged, the resulting transfer function would be identical. The system identification does not differentiate between the two-time constants.

\subsection{Single Down Pull Experiments}

\begin{figure}
    \centering
    \includegraphics[width = 0.5\textwidth]{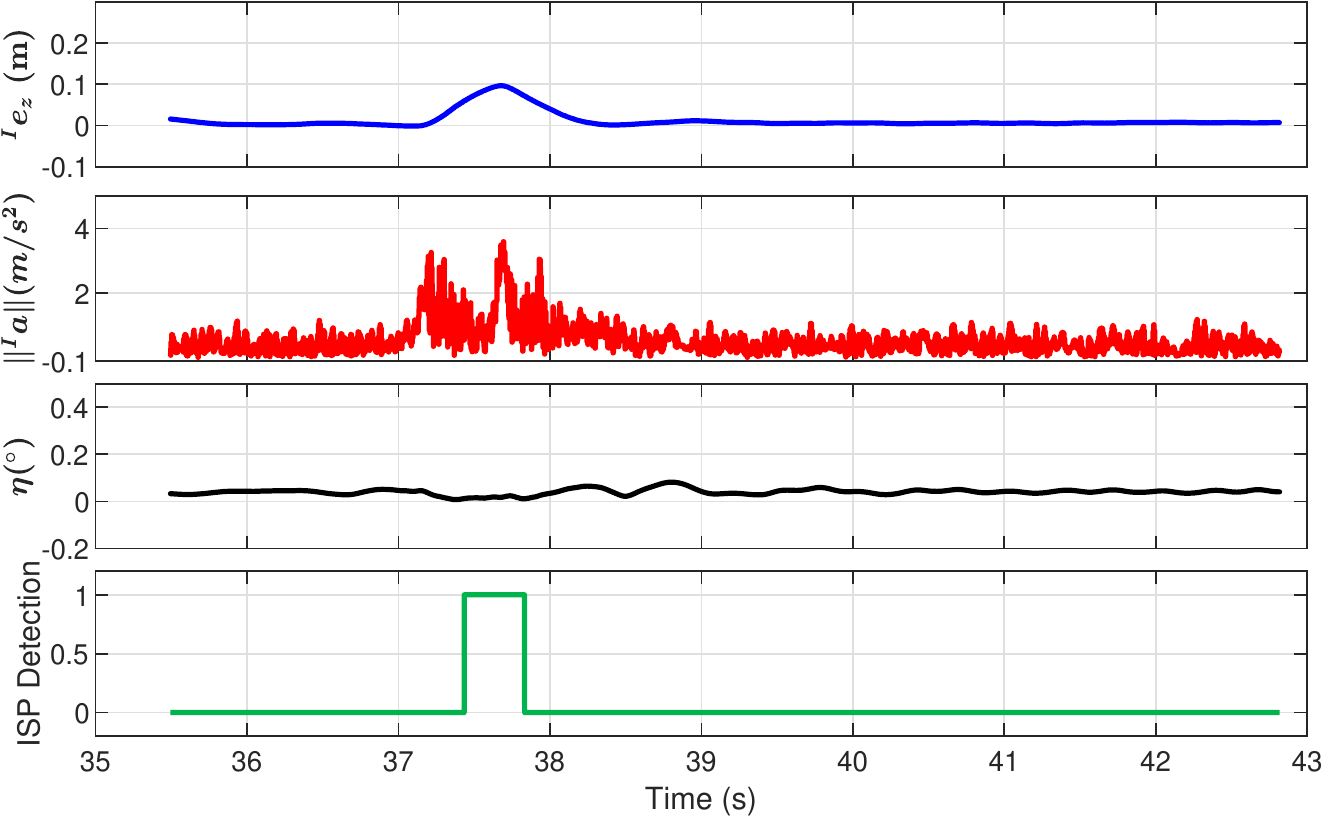}
    \caption{Detection of a single downward pull exerted on the base UAV}
    \label{fig:qdrone single pull}
\end{figure}

In this experiment, we test the ability of our method to detect an SDP trained and tested on the base UAV. To validate our method,
we perform 10 tests were the human intends to interact with the UAV (positive tests) and 40 tests where the forces are applied on the UAV without the intention to interact (negative tests). During the negative tests, random forces, with varying application time are applied on the UAV in random directions.
The confusion matrix of the above tests is shown in Table. \ref{table:confusion}, with an accuracy of 96\%.
Fig. \ref{fig:qdrone single pull} shows one example SDP experiment. This figure shows the modification in altitude of the platform $^Ie_z$, the platform acceleration norm $|| ^Ia ||$, yaw $\eta$, and the ISP detection flag.

\begin{table}[!h]
\centering
\begin{tabular}{|c|c|c|}
\hline
                & \textbf{predicted False} &\textbf{predicted True}  \\ \hline
                \textbf{actual False} & 40 & 0 \\ \hline
                \textbf{actual True} & 2 & 8 \\ \hline

\end{tabular}
\caption{confusion matrix for base UAV}
\label{table:confusion}
\end{table}

\subsection{ISP detection under wind disturbance and UAV motion}
\begin{figure}
    \centering
    \includegraphics[width = 0.5\textwidth]{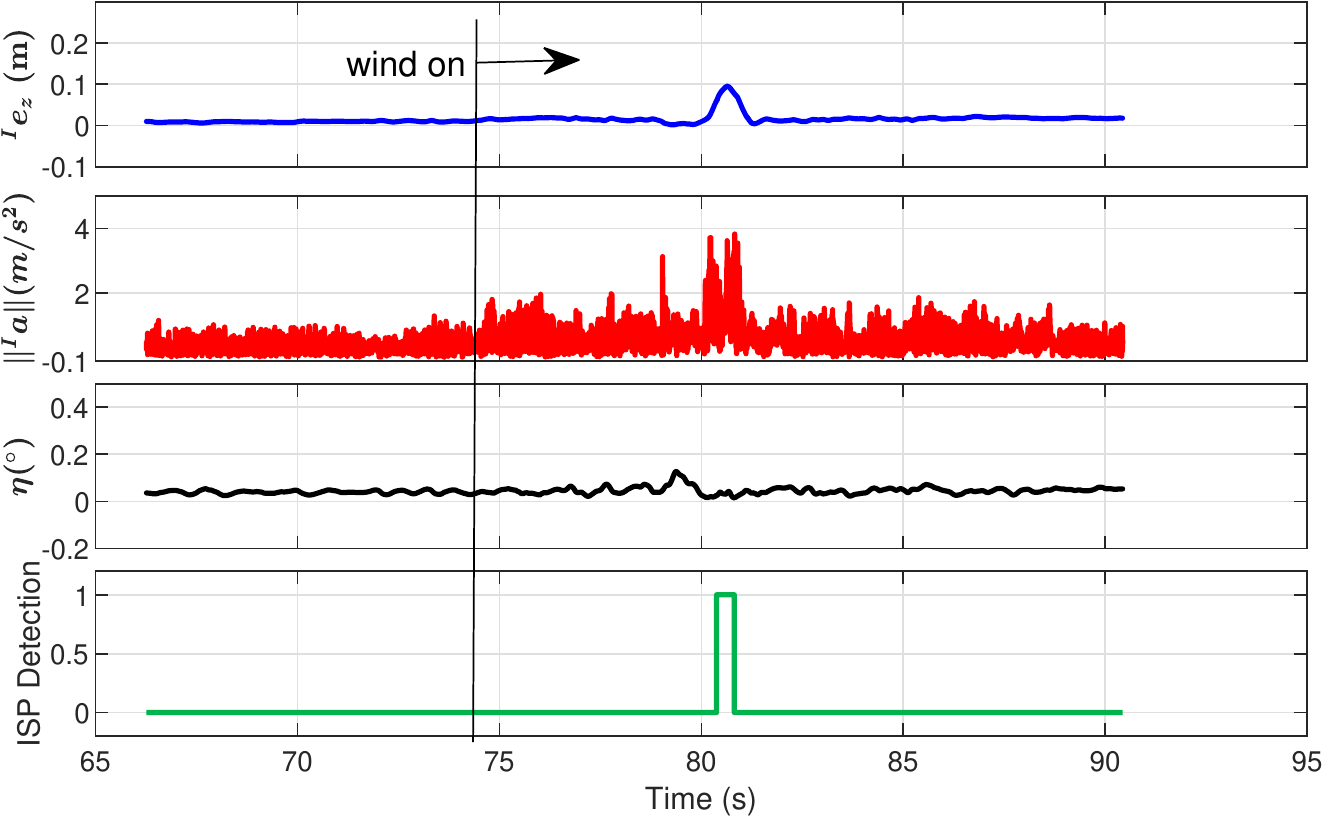}
    \caption{Detection of a single downward pull exerted on the base UAV under wind disturbance (at 5m/s)  }
    \label{fig:wind_disturbance}
\end{figure}

Once the ISP detection is validated on the base UAV, experiments are conducted to check the robustness of the trained detector to externally injected disturbances. A disturbance that is normally faced by UAVs is wind. To test the robustness of our approach to wind disturbances, we test the trained detector in the presence of lab generated wind, with speeds reaching $5m/s$. These experiments concluded that the proposed approach is able to distinguish wind disturbances from human interactions. Fig. \ref{fig:wind_disturbance} shows the deployment of our detector in the presence of wind, and demonstrates that the detector positively identifies the human interaction only despite the wind disturbance. 

\begin{figure}
    \centering
    \includegraphics[width = 0.5\textwidth]{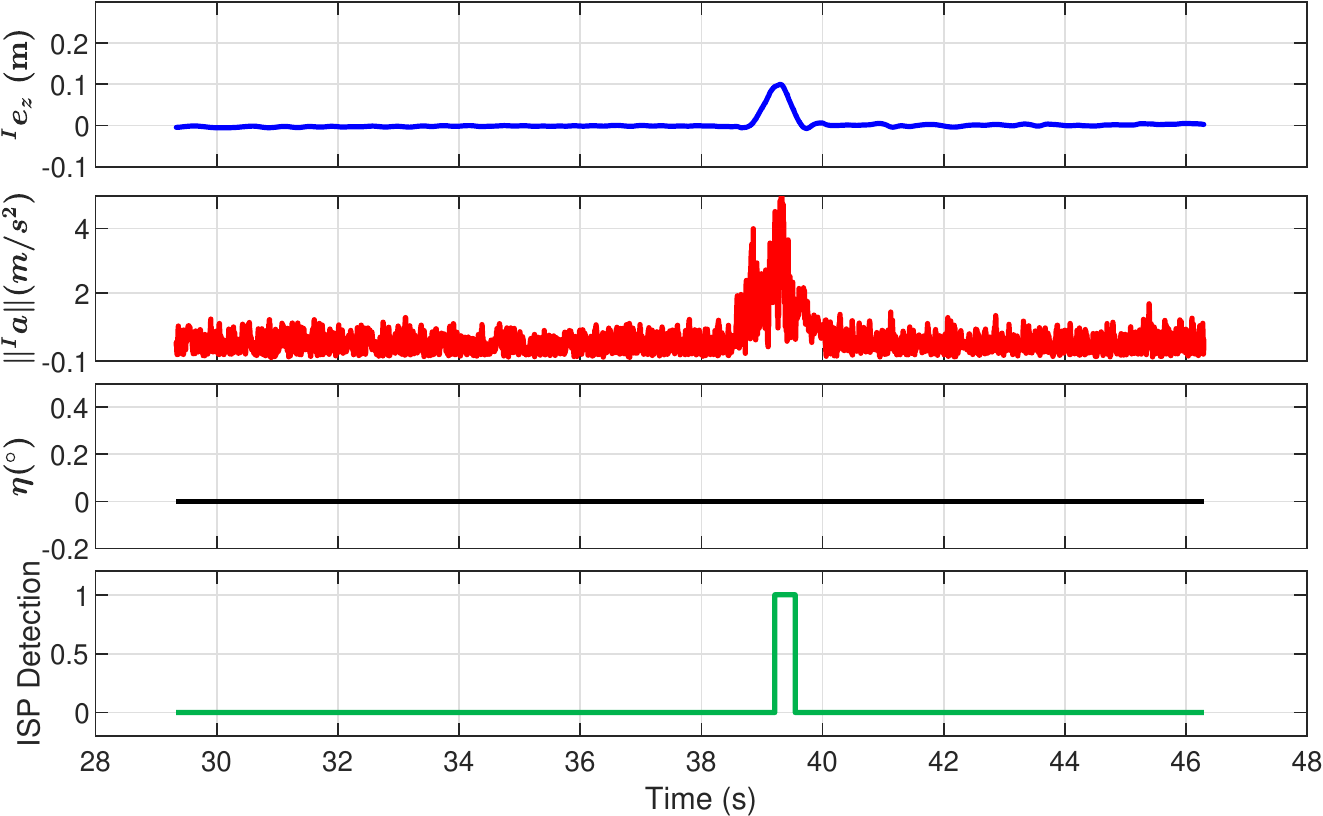}
    \caption{Detection of a single downward pull exerted on the base UAV while moving at 10cm/s}
    \label{fig:Moving_single_pull}
\end{figure}

In order to assess the detector's performance while the platform is in motion, an experiment is conducted where the UAV is commanded to move in the $y$ direction at a speed of 10cm/s; the speed was intentionally chosen small enough for the human to be able to interact with the platform in a safe manner. The detection of ISP while the UAV is moving is plotted in Fig. \ref{fig:Moving_single_pull}, showing the effectiveness of our detector in spite of the UAV motion.

\subsection{Detecting different ISPs}\label{other charecteristic}
To show that our detector learns to find force profiles instead of simple force thresholding, an experiment is designed where two consecutive pulls must be detected and lone pulls should be rejected, even though the direction of the pull is the same in both profiles.

\begin{figure*}[htpb]
     \centering
     \begin{subfigure}{0.45\textwidth}
         \vspace{15pt}
         \includegraphics [width=\textwidth]{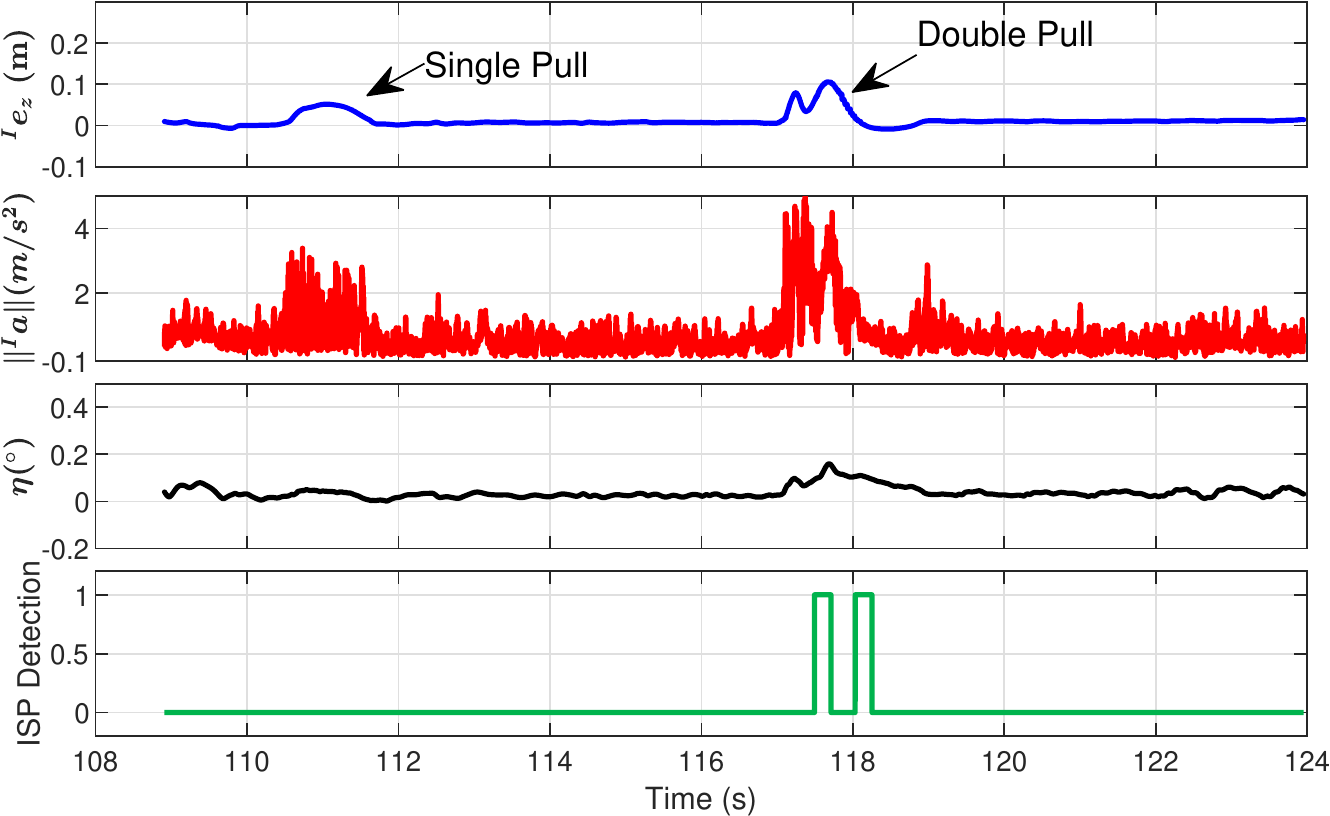}
         \label{fig:qdrone double pull detection}
     \end{subfigure}
     \hfill
     \begin{subfigure}{0.45\textwidth}
         \includegraphics[width=1\textwidth]{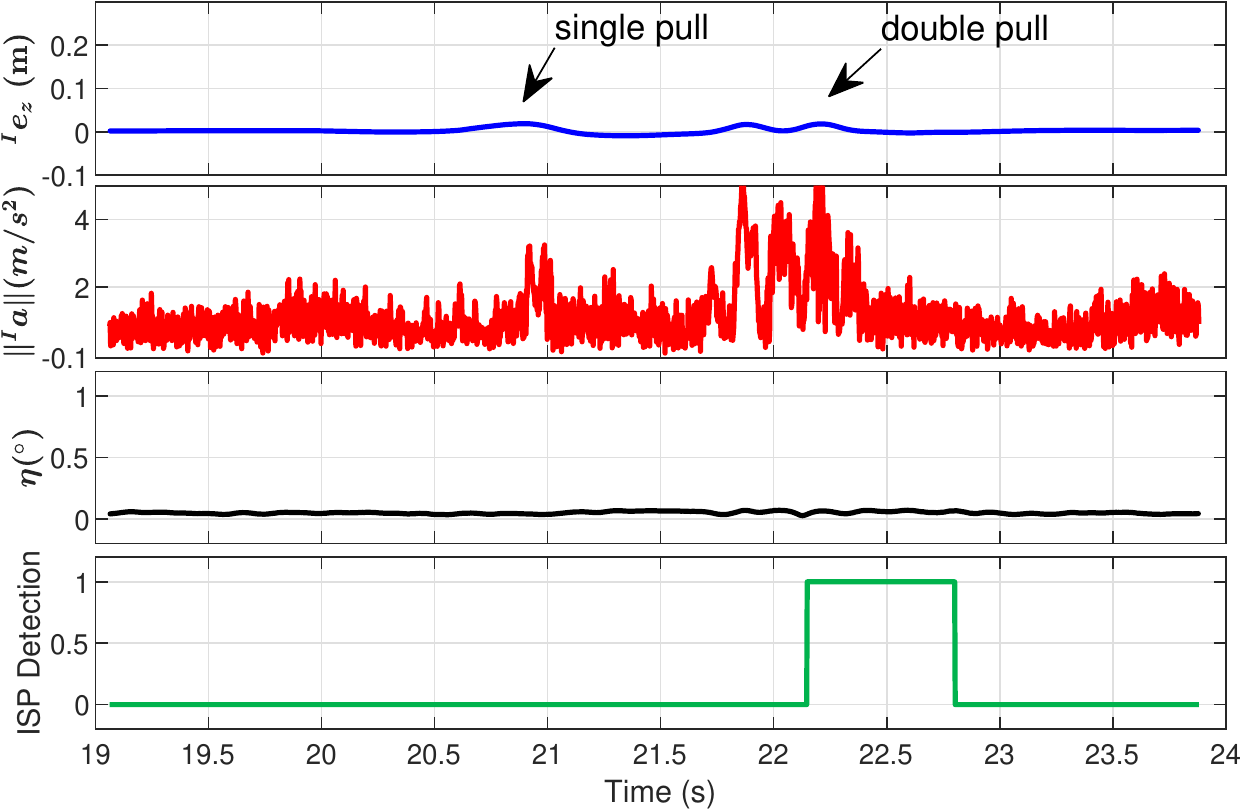}
     \end{subfigure}
     \caption{Detection of a consecutive downward double pull exerted on the base UAV (left plot), and the test UAV (right plot) using the same LSTM. The single pull was not detected in both cases, as expected}
     \label{fig:double pull}
\end{figure*}

In this experiment, our detector is trained to detect two consecutive pulls, and is tested on single and double pulls.
Fig. \ref{fig:double pull}(left) shows the results of this experiment. In this figure we can see the detector clearly identifying the double pull, while rejecting the single pull. 

Similar results were observed for the SYT ISP detection. We omitted the results of these experiments from this paper for brievity, however, these experiments are shown in the accompanying video \cite{paper_video}.

\subsection{Dynamics Tranferability}\label{hexa_validation}
As the CDDP ISP is the most complex between the three tested ISPs, we test the dynamic transformation between two different UAVs while detecting this ISP.
The testing UAV is only used to validate the dynamic transformation from the physical domain to the TID using the parameters from Table.~\ref{tab:hexa_params}. 
\begin{figure}[h]
    \centering
    \includegraphics[width = 0.5\textwidth]{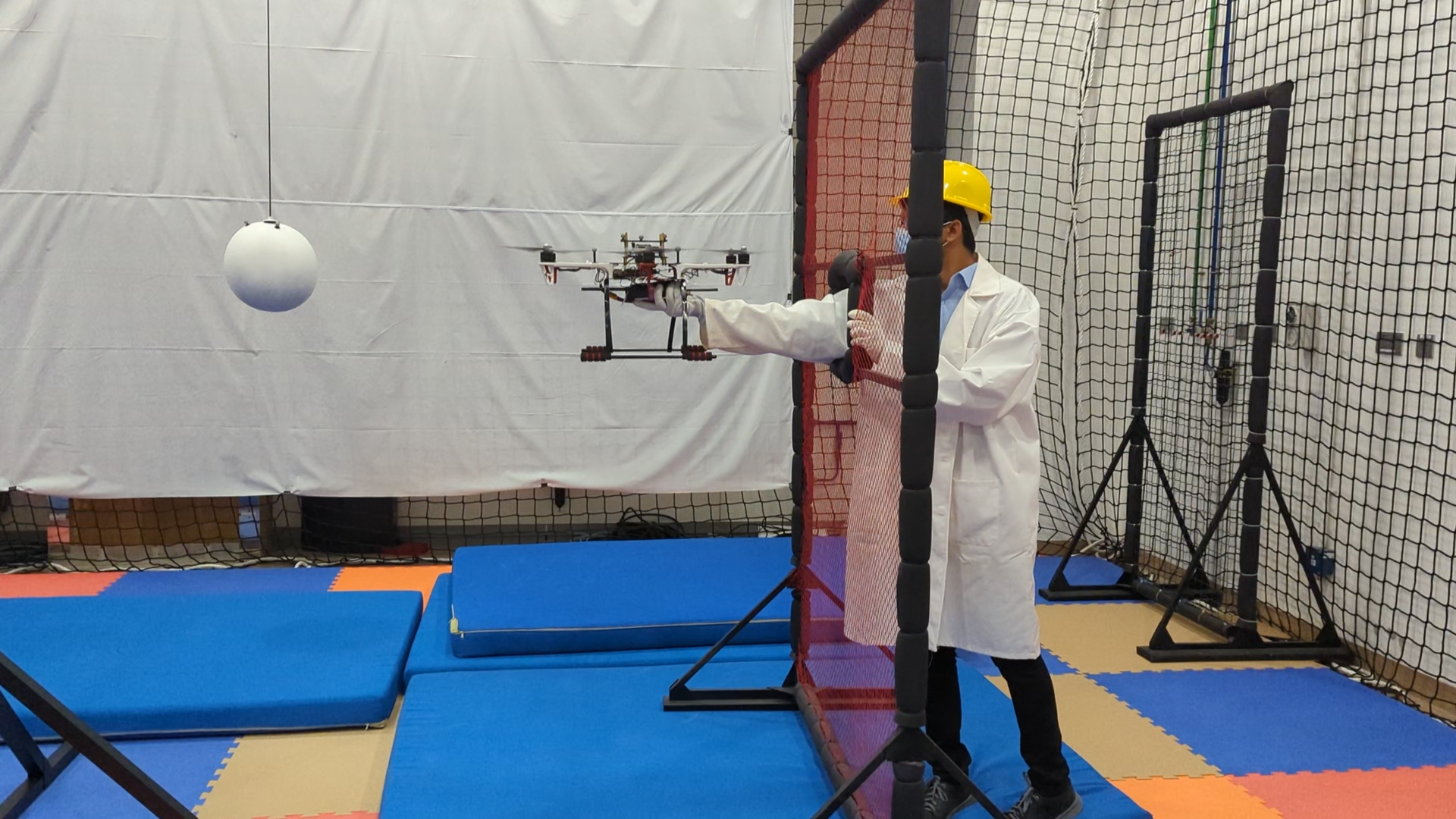}
    \caption{Interaction with the Test UAV in a safe and controlled setting}
    \label{fig:hexa_interaction}
\end{figure}

The results of this experiment are shown in Fig. \ref{fig:double pull}(right). From this figure, we can see that the human interaction is reliably detected on the testing UAV after the necessary dynamic scaling of the features.

It is worth noting that the same experiment was conducted with the testing UAV without any dynamic transformation. As expected, during this experiment, the testing UAV was not capable of detecting the human interaction.

\subsection{Payload Delivery} \label{payload delivery}
As an application to the ISP detection, we propose payload handover from a UAV to a human. The interaction detection system itself can be extended to other applications, however, this application is required in real-world payload delivery problems such as the ones in \cite{amazon, wing}, and can be used to validate our ISP detection method without any additional sensors or actuators that modify the platform's dynamics.

To offer a smooth user experience, once the UAV detects a human interaction, it should stop resisting the human and release it in a smooth manner. When an interaction is detected, the UAV is commanded to hold position for a set amount of time. This is done by changing the position reference to the estimated position at the start of the interaction.

\subsubsection{Gripper}\label{gripper}
\begin{figure}[h]
    \centering
    \includegraphics[width = 0.5\textwidth]{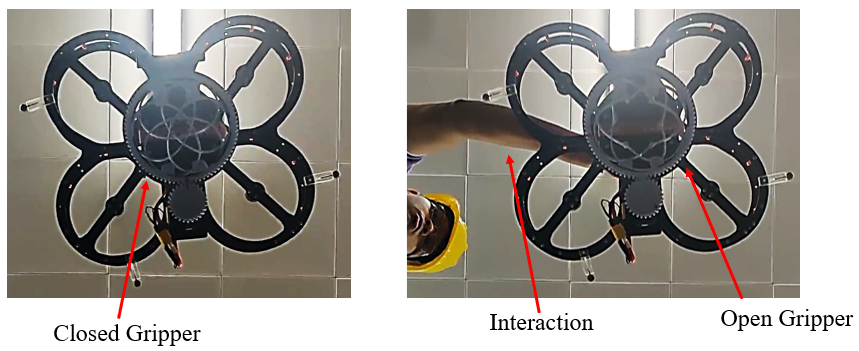}
    \caption{showing the designed girpper's orperation. The gripper is closed when no interaction is detected. When the human interacts with the UAV, the gripper opens to release the payload.}
    \label{fig:working_gripper}
\end{figure}

We designed a gripper based on the iris gripper
(\cite{flexsys}) that holds and releases the payload as required by the corresponding interaction.
%\begin{comment}
%\begin{figure}
%\centering
%    \includegraphics[width=0.4\textwidth]{images/iris gripper.png}
%        \caption{Iris Gripper}
%        \label{fig:IrisGripper}
%\end{figure}
%\end{comment}
The gripper is built using lightweight 3d printed material to allow maximum payload capacity, and is controlled with a servo motor(dynamixel Ax-12a). The gripper is firmly attached under the UAV.  The gripper can smoothly close to firmly grip the desired payload, and then, once an interaction is detected, it is commanded to open to release the payload. The gripper and the described operation is shown in Fig. \ref{fig:working_gripper}

%--------------------

\begin{comment}
\subsection{Robustness to Disturbance}
The Robustness of the proposed approach is tested by introducing disturbances. A common disturbance that is faced by UAVs during normal operation is that due to wind. We provide wind speeds not more than $5m/s$ to the UAV. It was found the that proposed approach is robust towards these wind speeds. This is shown in Fig. \ref{fig:windDetection}
\begin{figure}
    \centering
    \includegraphics[scale=0.35]{images/wind_disturbance plot-eps-converted-to.pdf}
    \caption{Human interaction detection in the presence of wind disturbances}
    \label{fig:windDetection}
\end{figure}

Another operating condition for most UAVs is while it is moving, The UAV is commanded to move in the y direction at a speed of 10cm/s. An interaction test was conducted while it is moving. It was found that the ISP is successfully detected in this case.

\end{comment}

 %------------------------------------------------

\section{Conclusion}\label{conclusion}

In this paper, a novel method for interaction detection between a human and a UAV is proposed. The presented method detects a state signature, referred to as the interaction states profile (ISP), exerted by the human on the UAV. The presented detection scheme is then demonstrated in a payload handover scenario, where following the ISP detection, the UAV releases the object for the human. The presented method is trained on a base UAV, with an LSTM-based neural network. The method is then transferred to other UAVs through a dynamics based transformation between the base UAV and the testing UAV, rendering the method agnostic to the training platform. The dynamics of the base and testing UAVs are exposed using the DNN-MRFT method~\cite{Ayyad2020}.
The presented approach is validated through an extensive experimental campaign, showing the detection of different ISPs at high success rate, the robustness of the presented method to wind disturbance, and the transferability between different platforms.

In the future, this work could be extended in different directions. For example, a bidirectional human-robot handover system can be developed to consider the case where a human hands an object over to a robot. In addition, the proposed approach can be extended to distinguish between multiple classes of force signatures simultaneously. Finally, while our method requires physical interaction between the human and the UAV to detect the interaction, the intention of the human to interact prior to the contact could be added to the handover scenario as was done in \cite{hamandi_cham_2018,hamandi_cham_2018_2}.

\medskip

%----------------------------------------------------------------------------------------
%	REFERENCE LIST
%----------------------------------------------------------------------------------------

\bibliographystyle{IEEEtran}
\bibliography{sources.bib}

\begin{IEEEbiography}[{\includegraphics[width=1in,height=1.25in,clip,keepaspectratio]{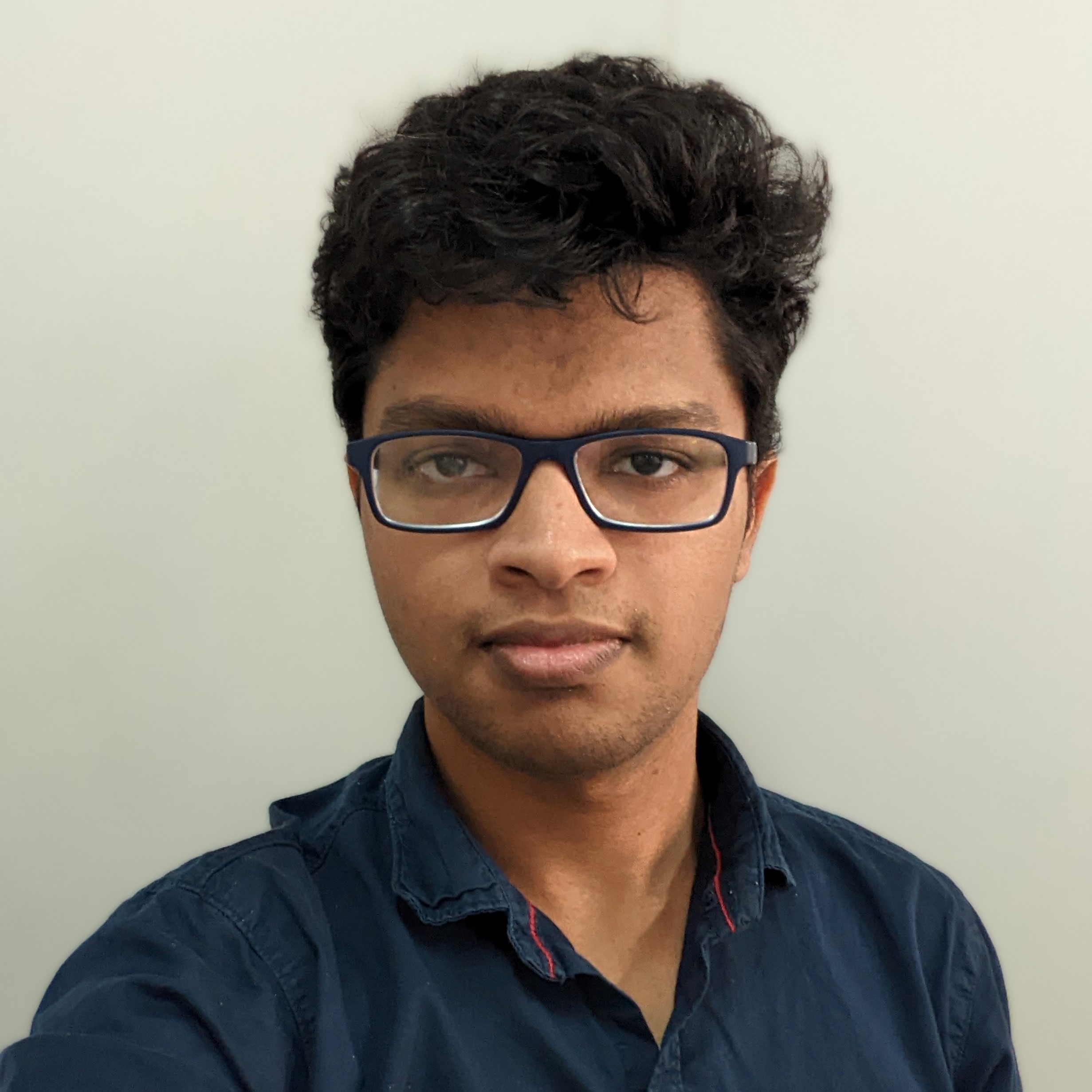}}]{Anees Peringal} recieved his BSc. degree in Aerospace engineering from Khalifa university, Abu Dhabi University. He is currently pursuing an MSc. degree in Aerospace engineering at Khalifa University. He is interested in research related to control of dynamic systems and autonomous robotics. 

\end{IEEEbiography}

\begin{IEEEbiography}[{\includegraphics[width=1in,height=1.25in,clip,keepaspectratio]{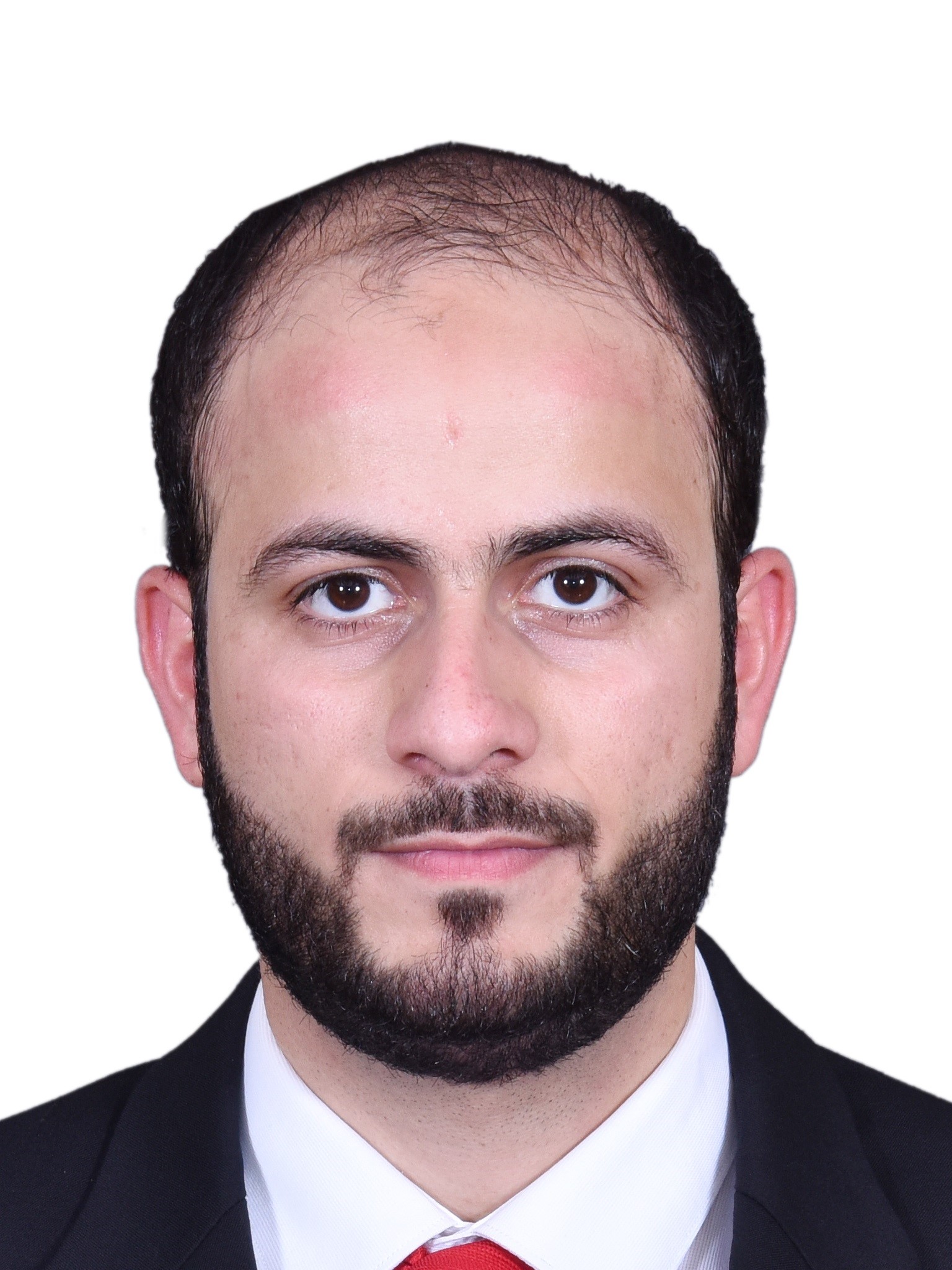}}]{Mohamad Chehadeh} received his MSc. in Electrical Engineering from Khalifa University, Abu Dhabi, UAE, in 2017. He is currently with Khalifa University Center for Autonomous Robotic Systems (KUCARS). His research interest is mainly focused on identification, perception, and control of complex dynamical systems utilizing the recent advancements in the field of AI.
\end{IEEEbiography}

\begin{IEEEbiography}[{\includegraphics[width=1in,height=1.1in,clip,keepaspectratio]{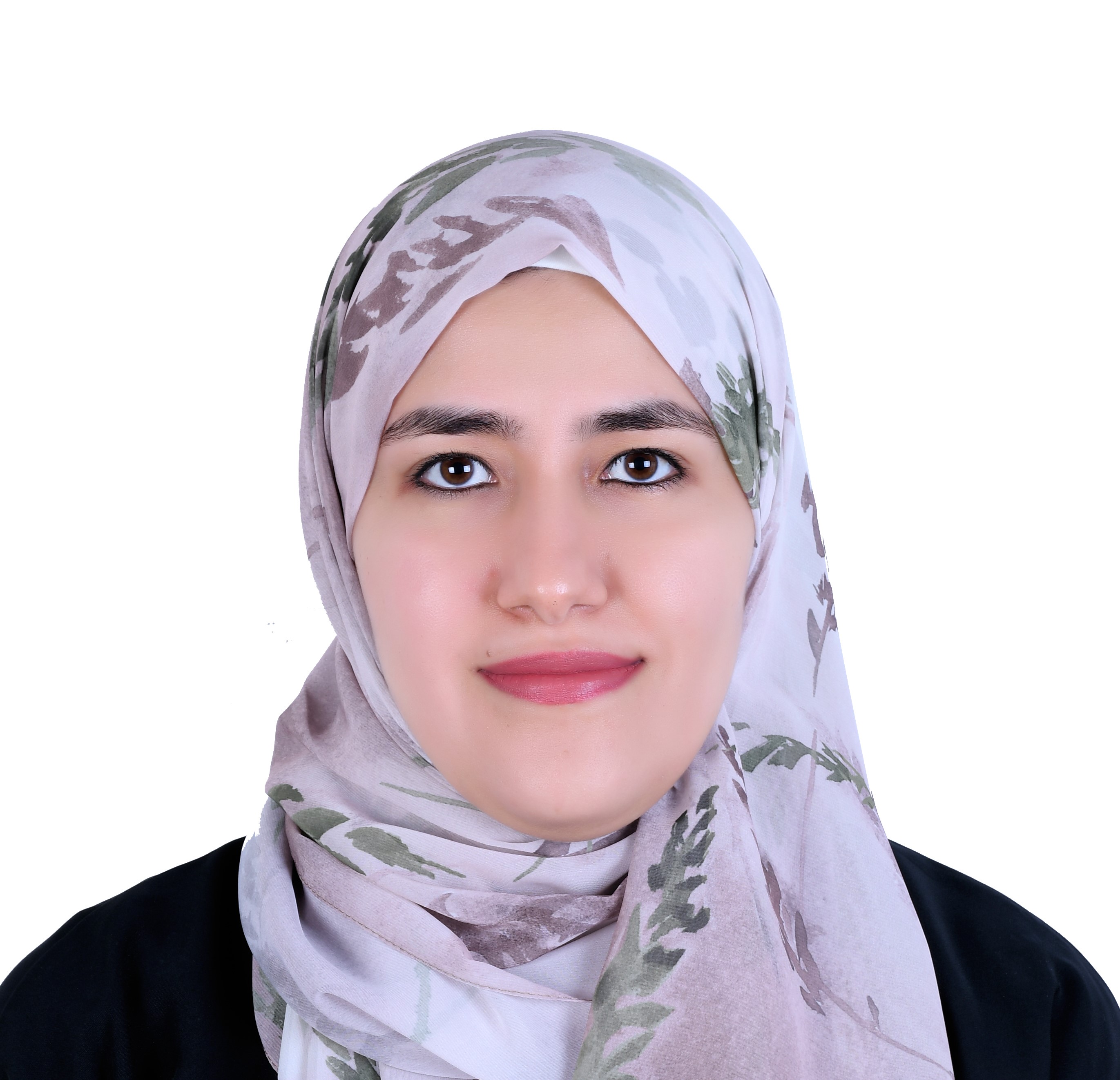}}]{Rana Azzam} received the B.Sc. degree in computer engineering and the M.Sc. degree by Research in electrical and computer engineering from Khalifa University in 2014 and 2016, respectively, and the Ph.D. degree in engineering with a focus on robotics in 2020. She is currently a Postdoctoral Fellow with the Department of Aerospace Engineering. Her research interests include machine learning, reinforcement learning, navigation, and simultaneous localization and mapping.
\end{IEEEbiography}

\begin{IEEEbiography}[{\includegraphics[width=1in,height=1.25in,clip,keepaspectratio]{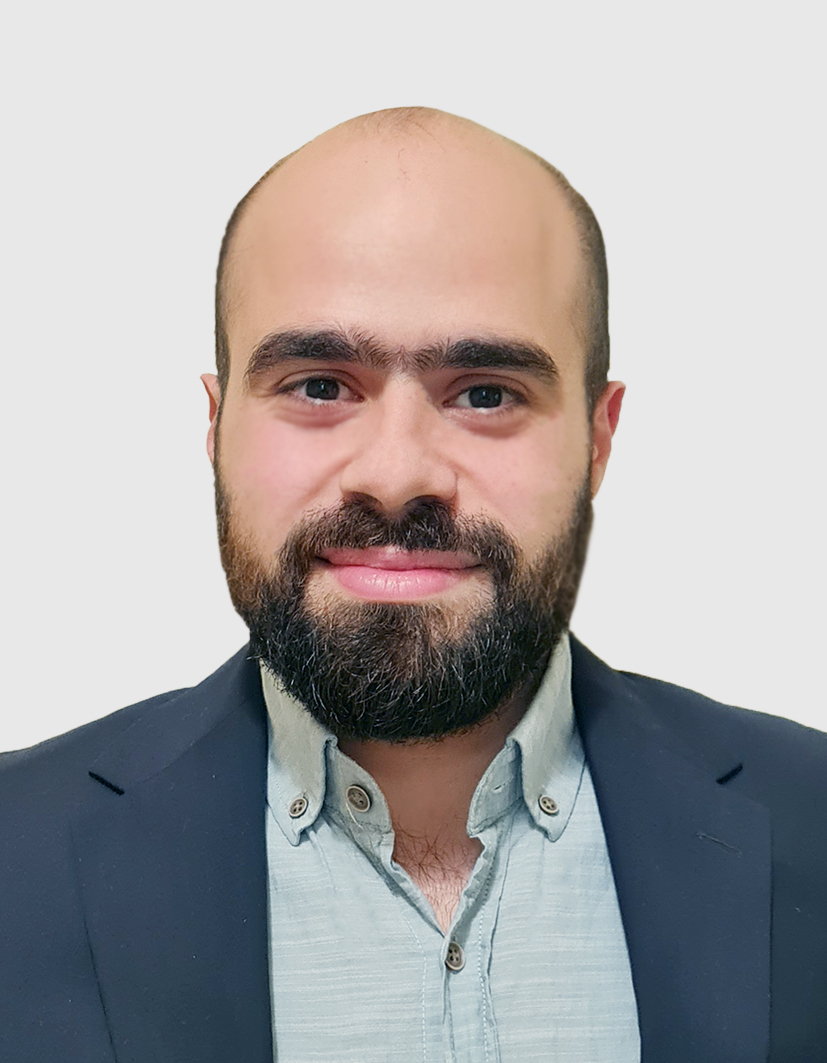}}]{Mahmoud Hamandi} received the MEng in Mechanical Engineering from the American University of Beirut, Beirut, Lebanon in 2017, and the Ph.D. in Automation and Robotics from the National Institute for Applied Sciences, Toulouse, France, in 2021. He is currently a Postdoctoral Fellow with Khalifa University Center for Autonomous Robotic Systems (KUCARS). His research interests include perception, design and control of robotic systems, with applications to aerial robots.
\end{IEEEbiography}

\begin{IEEEbiography}[{\includegraphics[width=1in,height=1.25in,clip,keepaspectratio]{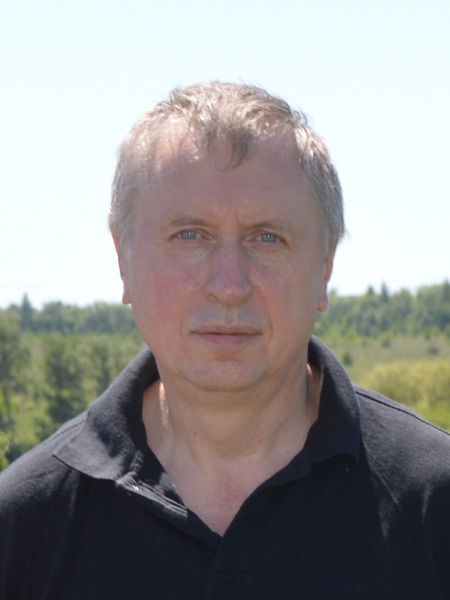}}]{Igor Boiko} received his MSc, PhD and DSc degrees
from Tula State University and Higher Attestation
Commission, Russia. His research interests include
frequency-domain methods of analysis and design of
nonlinear systems, discontinuous and sliding mode
control systems, PID control, process control theory
and applications. Currently he is a Professor with
Khalifa University, Abu Dhabi, UAE.
\end{IEEEbiography}

\begin{IEEEbiography}[{\includegraphics[width=1in,height=1.25in,clip,keepaspectratio]{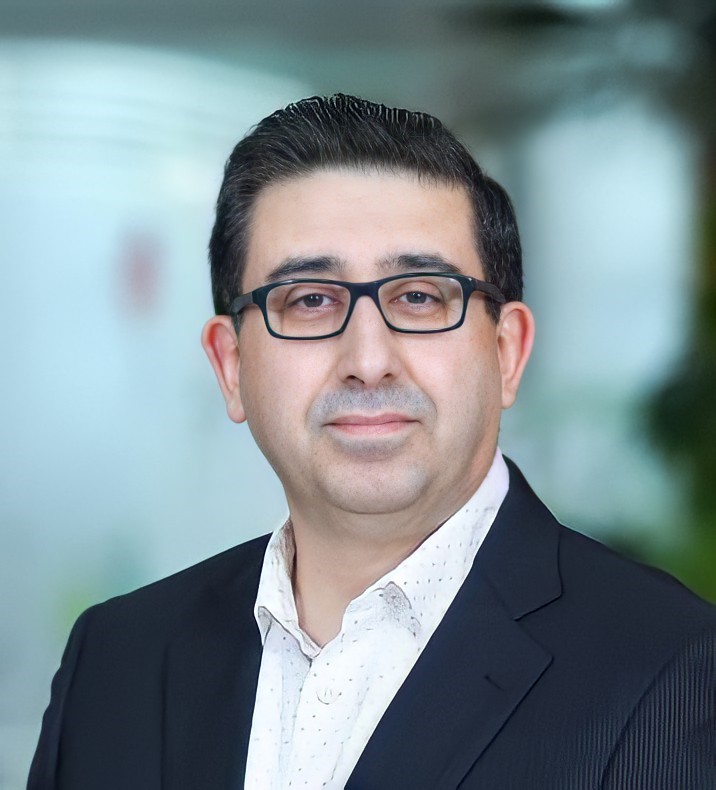}}]{Yahya Zweiri} (Member, IEEE) received the
Ph.D. degree from King's College London, in 2003. He is currently  an Associate Professor with the Department of Aerospace Engineering, and theme leader at  Khalifa University Center for Autonomous Robotic Systems (KUCARS), Khalifa University, United Arab Emirates. He was involved in defense and security research projects in the last 20 years at the Defence Science and Technology Laboratory, King's College London, and the King Abdullah II Design and Development Bureau, Jordan. He has published over 100 refereed journal and conference papers and filed ten patents in USA and U.K. in the unmanned systems field. His central research interests include interaction dynamics between unmanned systems and unknown environments by means of deep learning, machine intelligence, constrained optimization, and advanced control.
\end{IEEEbiography}

\EOD
\end{document}